\documentclass[10pt,journal,compsoc]{IEEEtran}

\ifCLASSOPTIONcompsoc
  \usepackage[nocompress]{cite}
\else
  \usepackage{cite}
\fi

\ifCLASSINFOpdf
\else
\fi

\usepackage{booktabs}       %
\usepackage{multirow}
\usepackage{amsmath,amssymb}
\usepackage{color}
\usepackage{caption}
\usepackage{subcaption}
\usepackage{graphicx}
\usepackage{xspace}
\usepackage{bbm}
\usepackage{makecell}
\usepackage{gensymb}
\usepackage{hyperref}

\usepackage{fontawesome}
\usepackage{xcolor}
\definecolor{dartmouthgreen}{rgb}{0.05, 0.5, 0.06}

\newtheorem{definition}{Definition}

\newcommand{\mc}{\mathcal}
\newcommand{\R}{\mathbb{R}}
\newcommand{\E}{\mathbb{E}}
\newcommand{\D}{\mathcal{D}}

\newcommand{\norm}[1]{\left|\left|#1\right|\right|}

\makeatletter
\DeclareRobustCommand\onedot{\futurelet\@let@token\@onedot}

\DeclareMathOperator*{\argmax}{arg\,max}
\DeclareMathOperator*{\argmin}{arg\,min}

\def\@onedot{\ifx\@let@token.\else.\null\fi\xspace}

\def\eg{\emph{e.g}\onedot} 
\def\ie{\emph{i.e}\onedot} 
 
\def\etc{\emph{etc}\onedot} 
 
\def\etal{\emph{et al}\onedot}
\usepackage{enumitem}
\begin{document}
\title{Learning Compositional Representations for Effective Low-Shot Generalization}

\author{Samarth Mishra*,
        Pengkai Zhu*,
        and~Venkatesh~Saligrama,~\IEEEmembership{Fellow,~IEEE} %
\IEEEcompsocitemizethanks{
\IEEEcompsocthanksitem *: Equal contribution.
\IEEEcompsocthanksitem Pengkai and Venkatesh are with the Department of Electrical and Computer Engineering at Boston University. \protect\\
Primary contact: zpk@bu.edu
\IEEEcompsocthanksitem Samarth is with the Department of Computer Science at Boston University.
}%
}

\markboth{Submitted to IEEE TRANSACTIONS ON PATTERN ANALYSIS AND MACHINE INTELLIGENCE, Apr 2021}%
{Shell \MakeLowercase{\textit{et al.}}: Bare Demo of IEEEtran.cls for Computer Society Journals}

\IEEEtitleabstractindextext{%
\begin{abstract}
    We propose Recognition as Part Composition (RPC), an image encoding approach inspired by human cognition. It is based on the cognitive theory that humans recognize complex objects by components, and that they build a small compact vocabulary of concepts to represent each instance with. RPC encodes images by first decomposing them into salient parts, and then encoding each part as a mixture of a small number of prototypes, each representing a certain concept. We find that this type of learning inspired by human cognition can overcome hurdles faced by deep convolutional networks in low-shot generalization tasks, like zero-shot learning, few-shot learning and unsupervised domain adaptation. Furthermore, we find a classifier using an RPC image encoder is fairly robust to adversarial attacks, that deep neural networks are known to be prone to. Given that our image encoding principle is based on human cognition, one would expect the encodings to be interpretable by humans, which we find to be the case via crowd-sourcing experiments. Finally, we propose an application of these interpretable encodings in the form of generating synthetic attribute annotations for evaluating zero-shot learning methods on new datasets.
\end{abstract}

}

\maketitle

\IEEEdisplaynontitleabstractindextext

\IEEEpeerreviewmaketitle

\IEEEraisesectionheading{\section{Introduction}\label{sec:introduction}}

\IEEEPARstart{D}{eep} convolutional networks (DCNs) although effective at image classification, need large amounts of annotated images to learn from \cite{krizhevsky2012imagenet}. They encounter major hurdles when learning from a few labeled examples \cite{vinyals2016matching, ravi2016optimization}. Moreover, the image features that these networks learn are often quite fickle and do not adapt to changes in the image domain and are also susceptible to adversarial image perturbations \cite{szegedy2013intriguing}, imperceptibly small to the human eye.

Humans, even children, on the other hand, do not face these challenges and can effectively learn concepts having only seen a few examples of it \cite{benelli1991markman, feldman1997structure, xu2007word}. A well accepted theory behind this capability of humans is the ability to ``recognize by components'' \cite{biederman1987recognition}. In other words humans can learn to recognize concepts as a composition of simpler pieces. Lake~\etal~\cite{lake2015human} attempted at leveraging this in a Bayesian Program Learning framework, to build a system for effective one-shot generalization at tasks like recognizing categories of hand-written characters, and found low-shot generalization comparable to that of humans and much better than state of the art DCNs for image recognition.

\begin{figure*}
    \centering
    \includegraphics[width=\linewidth]{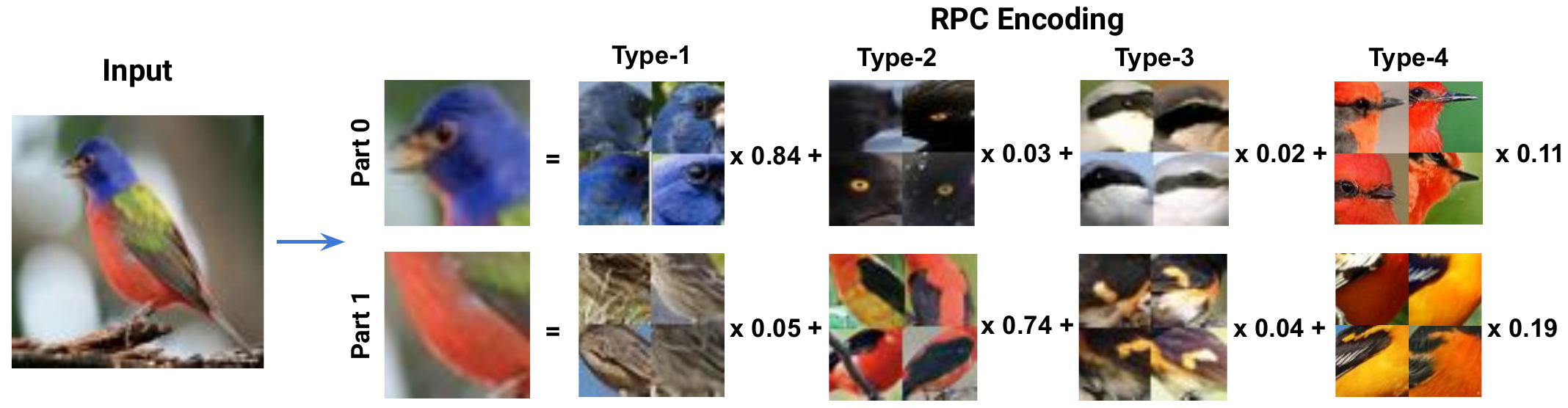}
    \vspace{-0.5cm}
    \caption{Recognition as Part Composition (RPC) learns to recognize representative parts in an image and represents each part instance using a mixture of prototypes, each representing a concept/part-type. The coefficients of the mixture are the likelihoods that that specific instance is of the corresponding part-type. 
    }
    \label{fig:embedding}
\end{figure*}

Decomposing an image into parts is a recognition approach suggested by this theory. Additionally, we draw inspiration from quantum cognition theory, that suggests each instance of an object typically is learnt as a superposition of multiple concepts \cite{quantum_cognition}. An instance can be represented as a weighted sum of these concepts with coefficients representing collapse probabilities. Under this framework, each part recognized in an image can be represented as a convex combination of a vocabulary of few concepts or prototypes. 

Inspired from this theory, we propose Recognition as Part Composition (RPC), a method that learns image representations by first decomposing it into a few semantically representative parts, and then learning an encoding of each part in terms of a small number of prototypes. For discovering different parts of an image, we use a multi-attention convolutional neural network (MACNN) (similar to the model used by \cite{zheng2017learning}), and unlike the approach of Lake~\etal~\cite{lake2015human}, which used additional annotations in the form of pen strokes for character recognition, our model automatically recognizes representative parts of an image. Only annotations we need to use for our model are class-labels for images.

To illustrate the encodings generated by RPC, we show an example in Fig \ref{fig:embedding}. Our model learns to recognize key parts of a bird like its head (part-0), breast (part-1) \etc. Additionally it learns certain prototypical types for each part, which we represent via 4 images of parts closest to that type in Fig. \ref{fig:embedding}. Our model produces a distribution over part-types representing the likelihood that the given part comes from a certain part-type, for each part recognized in the image. The RPC encoding is a collection of these distributions for each part recognized by our model. This image encoding has surprisingly good low-shot generalization properties as we shall see in our evaluations in few-shot learning, zero-shot learning and domain adaptive image recognition tasks. This indicates that concepts learnt by our method using a training set of images, generalize well to an unseen set. We also find that these representations are interpretable by humans and can be robust to adversarial image perturbations. 

Explainability is favored in models that are deployed for making predictions in the real world, so human operators can trust the model's decision and possibly diagnose problems when they arise. DCNs, however, with their high dimensional image representations and highly non-linear reasoning end up becoming ``black-box'' models, making it difficult to understand the reasoning behind their decisions. Conventional wisdom suggests that representations learned end-to-end for a task achieve better performance than semantic representations designed to be interpretable. Thus existing efforts mostly focus on explaining end-to-end networks post-hoc, by grounding their decision in image pixels via an attention map or active patches~\cite{selvaraju2017grad,petsiuk2018rise,bargal2018excitation}. In contrast, we learn explainable representations in an end-to-end manner, and our method enhances rather than limits the system's ability to efficiently learn low-shot concepts and transfer to novel visual domains and small-data problems.

This paper builds on our preliminary work \cite{zhu2019learning}, and extends it substantially in multiple directions, exploring new concepts and applications to novel scenarios. We summarize these contributions below:
\begin{itemize}[leftmargin=8pt,nosep]
    \item {\it Compositionality of Concepts.} We present Recognition as Part Composition (RPC), a general image encoder that is inspired by human cognition and produces image encodings by decomposing an image into parts and then representing each part in compact vocabulary of a few concepts.
    \item {\it Low-Shot Generalization.} We show that the RPC encoding is useful in low-shot generalization tasks like few-shot learning, zero-shot learning and visual domain adaptation, and that a simple model using an RPC encoder performs favorably compared to state of the art methods in those tasks. 
    \item {\it Robustness to Adversarial Attacks.}
    A classifier with an RPC encoder can be robust to {\it adversarial} attacks, than a standard DCN. This is because the RPC encoder embeds inputs into a space of discrete concepts, and thus for an attack to be successful, the adversary must modify the input significantly to modify a concept.
    \item {\it Interpretability.} We also show that RPC encodings are also interpretable by humans by crowd-sourcing questions regarding our model encodings.
    \item {\it Synthesizing Datasets.} Finally, given that RPC encodings are human interpretable we propose another possible application of our model as a synthetic attribute generator for evaluating zero-shot learning methods on new datasets.
\end{itemize}

We note while mentioning our contributions that the goal of this paper is an exposition of certain nice properties our RPC encodings have. While on different low-shot generalization tasks, we will see that a model using these image representations performs competitively with recent approaches specifically developed for those tasks, the goal is not to show RPC achieves performance better than those task-specific approaches. Rather, our framework should be viewed as exposing the benefits of human-like learning for limited-shot learning problems.

\section{Related Work} \label{sec:related_work}

In the introduction, we discussed how compositionality and grounding objects in a vocabulary of a few simple concepts are key ingredients in human intelligence. Lake~\etal~\cite{lake2015human} followed this via a somewhat restrictive model of Bayesian Program Learning. In contrast, our RPC model, while leveraging recent advances in deep neural networks, induces desirable properties that are evident in our understanding of human intelligence. In the sequel, we discuss prior work in three different domains. Note that with the range of research that has gone into each of these problems, this section provides only a flavor of related research and readers are suggested to read review articles on the topics for a more thorough and wider coverage of them (\eg \cite{xu2020adversarial, vilone2020explainable}). 

\textbf{Explainability.} As mentioned in the introduction a large part of recent explainability research is driven by post-hoc analysis of deep networks. This is because these models although highly non-linear with high dimensional feature spaces, have been the strongest performers on recognition tasks. Many approaches search for prediction explanations in the form of saliency maps. Among them, a common approach is to use gradient magnitudes for different class activations \cite{selvaraju2017grad, zhou2016learning, bargal2018excitation}. While this assumes access to the model and hence its gradients, some other approaches like \cite{petsiuk2018rise} aim to get a saliency map by generating random masks for input images and using the network prediction to determine the closeness of the random mask to the actual saliency map. Another set of approaches tries to assign semantic concepts to hidden neurons in a neural network \cite{bau2017network, bau2020understanding}. Our approach is different from these post-hoc interpretability analyses in that our image encodings are a composition of a small number of concepts that are interpretable by humans, as we shall see in our experiments. We find that the restrictions that a small number of parts and concepts can create, do not encumber our model's low-shot generalization capabilities, but rather enhance it.

\noindent \textbf{Adversarial attacks and robustness.} Szegedy~\etal~\cite{szegedy2013intriguing} first discovered that deep neural networks are vulnerable to adversarial examples, which are only slightly different to correctly classified examples from the original data distribution. The examples on the Imagenet dataset \cite{deng2009imagenet} were often so close to the original examples that the two were indistinguishable to humans. Subsequent research led to the development of different kinds of adversarial attacks on images, both in the white-box scenario \cite{goodfellow2014explaining, kurakin2016adversarial, papernot2016limitations, moosavi2016deepfool} where adversaries can access the trained model and the black-box scenario \cite{papernot2017practical, chen2017zoo}, where they cannot. Various defense strategies against these attacks have also been developed \cite{papernot2016distillation, buckman2018thermometer, dhillon2018stochastic}. A simple white box attack that was found to reliably generate adversarial examples for a wide range of models was the fast gradient sign method (FGSM) \cite{goodfellow2014explaining}, and they proposed training of the deep neural networks with such adversarial examples as one possible strategy for defending against them. We evaluated our RPC model using adversarial examples from such an FGSM attack and found it to have much better robustness to them as compared to a deep convolutional network trained on the same task. Note that no adversarial examples were used in the training of the two models.

\noindent \textbf{Low shot generalization. } We use this term to refer to a collection of problems with some constraints surrounding availability of labeled data from the target distribution. Note that most of these problems do assume an abundance of some form of training data, but the scarcity arises because the model is tested on its predictions made on data from a different target distribution than the one it was trained on. Information about this target distribution is unavailable or scarcely available to the model during training. We discuss 3 such problems in recognition tasks: \\
(1) {\it Unsupervised Domain Adaptation (UDA).} This problem focuses on a task where at inference time, images come from a target distribution or domain that has the same set of classes, but has visually dissimilar images to the ones in the training or source distribution. We specifically focus on the scenario where the model has access to unlabeled images from the target distribution. Ben-David~\etal~\cite{ben2010theory} provided an upper bound on a classifier's target error consisting of its source error and a divergence between the two distributions. Using this result many approaches were derived that attempt to solve UDA by aligning the two distributions in feature space \cite{ganin2016domain,tzeng2017adversarial,long2018conditional,saito2017adversarial, saito2018maximum}. Some other approaches like \cite{saito2017asymmetric, chadha2018improving} leverage the transductivity of the problem to use pseudo-labeling for unlabeled target domain images. Still others attempt to train generative models to translate source images to target images allowing to train a classifier with these generated target labeled images \cite{hoffman2017cycada, taigman2016unsupervised, shrivastava2017learning, bousmalis2017unsupervised}. In our evaluation of RPC, we utilize a pseudo-labeling approach that is a simpler version of \cite{saito2017asymmetric}.  \\
(2) {\it Few-Shot Learning (FSL).} This problem tests a model's ability to learn classification using a few labeled examples per class. Typically the model is allowed to be trained on a label abundant training dataset, and at inference time, it ``adapts'' to the classification problem defined by the few labeled examples. Some common approaches here have attempted to learn a feature space, with a class-defined neighborhood in a distance metric (say $\ell_2$). This allows for a non-parametric ``adaptation'' at inference time, using simply a nearest neighbors classifier in the feature space \cite{vinyals2016matching, snell2017prototypical, sung2018learning}. Meta-learning or learning-to-learn (or more aptly for this problem, learning-to-adapt) is an approach also quite quite popular \cite{finn2017model, ravi2016optimization, munkhdalai2017meta}. Finally, generative models like GANs \cite{goodfellow2014generative} have also found use in this problem, with approaches augmenting the few-labeled examples with generated ones for classifier learning \cite{antoniou2017data, wang2018low, mehrotra2017generative}. For our RPC model, we do nearest neighbor classification at inference time similar to \cite{snell2017prototypical}, but we train the model for classification on the training set using a 2 layer MLP classifier on the RPC encodings. \\
(3) {\it Zero-Shot Learning (GZSL).} This problem, at inference time, requires a model to make class predictions on images where certain classes (simply termed ``unseen classes'') have no examples in the training set. The classifier accesses some semantic vector representation of classes, to relate an example of an unseen class to its semantic vector. Some common approaches include \cite{frome2013devise,Lee_2018_CVPR,Wang_2018_CVPR} which learn feature embeddings that directly map the visual domain to the semantic domain and infer classifiers for unseen classes. Some other approaches like \cite{sung2018learning} learn a distance metric between the images and the semantic vectors of the class they belong to. Similar to both the above problems, some approaches have also attempted to synthesize unseen class images in the new environment from the given semantic attributes \cite{zhu2018generative,Verma_2018_CVPR,Xian_2018_CVPR,Jiang_2018_ECCV}. We trained the RPC model to learn a simple distance metric between the image feature space and the semantic vectors for this problem.

\noindent {\bf Recognition using object parts} has been an ingredient in multiple recognition approaches over the years. Ullman~\etal~\cite{ullman2002visual}, showed that information maximization with respect to classes of images resulted in visual features that were of intermediate complexity (neither too high, nor too low), indicating these were optimally beneficial for classification. These would result in selecting components like eyes, mouth, \etc in facial images and tyres, bumper, windows \etc in images of cars, further motivating recognition by components. Object geometries were paid particular attention to in Deformable Part Models \cite{felzenszwalb2009object, felzenszwalb2010cascade} which learned class models using both part features and their geometries, and assigned probability scores for object presence based on both how well the features matched as well as how much the geometry was deformed. Originally the method used Histogram of Gradients (HoG) features, but subequently, DPMs were also developed with deep CNNs ~\cite{savalle2014deformable,girshick2015deformable}. In fine-grained recognition multi-attention has been used in prior works for recognizing different object parts in images \cite{zheng2017learning}. Attention based approaches \cite{locatello2020object, greff2019multi, burgess2019monet} have also been found to be able to discover different semantically meaningful parts like objects and background components in images of synthetically generated 3D scenes. RPC uses attention maps corresponding to different object parts as well, along with specific priors that incorporate properties which make the RPC encoding interpretable (described in sections \ref{sec:method} and \ref{sec:explainability}).

\section{Recognition as Part Composition} \label{sec:method}

Our approach learns an image classifier by learning an intermediate feature representation that we refer to simply as the RPC encoding. We describe the notation used in our exposition: $x \in \mc{X} \subset \R^{D}$ denotes an image and $y \in \mc{Y}$ denotes a class label for the image. We will use $p(x, y)$ to denote the joint distribution over image, label pairs.

\noindent \textbf{Inductive Bias.} Our proposed approach injects an inductive bias, to factorize a task posterior $p_t(y|x)$ into a task agnostic mapping function, $\pi : \mc{X} \rightarrow \Pi$ or the RPC encoder, and a task-specific posterior acting on $\pi(x)$ for $x \in \mc{X}$. The mapping $\pi$ involves learnable parameters as we shall see in Sections \ref{subsec:part-feat} and \ref{subsec:ptle}. Thus \footnote{Note that although we make these statements for the true task distribution, we can only impose the bias in the predicted posterior. RPC's generalization power leads us to believe that the true posterior for our evaluation tasks may follow a similar factorization.}
\begin{align}
    p_t(y|x) = p_t(y|\pi(x))
\end{align}

The above structural decompostion is useful only if the latent variable $\pi(x)$ which is a task independent embedding, does the ``heavy-lifting'' allowing to distill predictive information from the training distribution to the target task distribution. Note that the decomposition itself is not special and it is not clear whether we can learn maps that allow for such information transfer. For instance, if $\pi$ is the identity map, no information is distilled from the training task distribution for the target tasks. We however, draw on insights from human cognition to propose our RPC encoder, which, as we shall see from our experiments, does allow for effective knowledge transfer from training tasks.

\begin{figure*}
\centering
\includegraphics[width=0.95\textwidth]{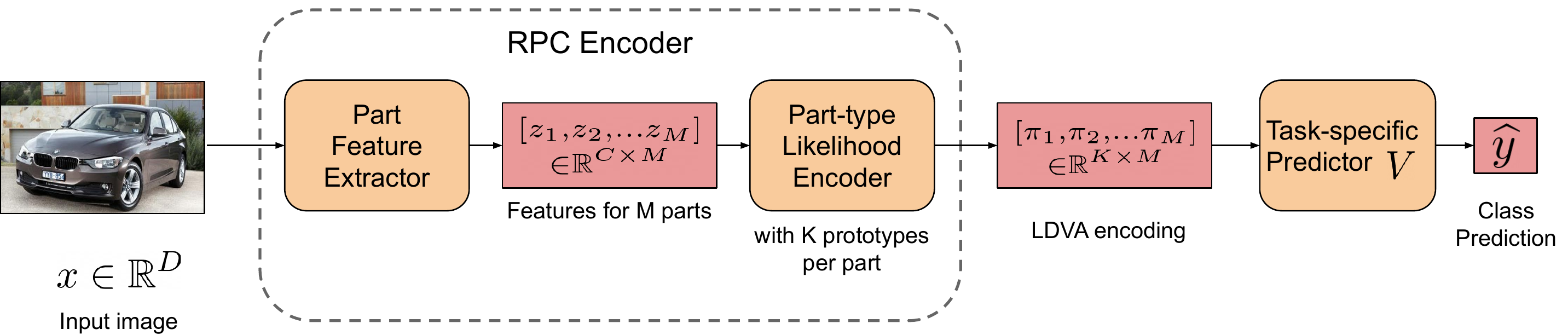}
\caption{\textbf{Recognition as Part Composition (RPC).} For an input image $x$, the part feature extractor decomposes $x$ into $M$ parts and extracts associated features $z_m$. The Part-type Likelihood encoder then encodes each part feature as a low-dimensional encoding $\pi_m$ by projecting the features onto a dictionary of part prototypes automatically discovered by the RPC model. $\pi_m$ is then used as input to train task-specific predictor models for GZSL, FSL and UDA. The model is trained end-to-end for each task}
\label{fig:model}
\end{figure*}

An overview of the model is shown in Fig \ref{fig:model}. Given an image $x \in \mc{X}$, the model first identifies $M$ parts in the image and extracts features $z$ for each part using the ``Part feature extractor'' (Sec. \ref{subsec:part-feat}). It then uses the Part-type likehood encoder (Sec. \ref{subsec:ptle}), which learns certain prototypical representations for the different part types, to generate an encoding for each part as a convex combination of these prototypes. These features denoted using $\pi$ is the RPC encoding of the image. Finally, depending on the task, a different task specific predictor $V$ (Sec. \ref{subsec:task-specific}) is chosen, which outputs a class prediction $\widehat{y} = V(\pi(x))$ for the image $x$, and the model is trained end-to-end.

\subsection{Part Feature Extractor} \label{subsec:part-feat}
Inspired by \cite{zheng2017learning}, we use a multi-attention convolutional neural network (MACNN) to map input images into a finite set of part feature vectors, $z_m \in \mathbb{R}^C$. It contains a module $E$ which, for input image $x$ generates a convolutional feature representation $E(x) \in \R^{W \times H \times C}$ with width $W$, height $H$ and $C$ channels. A second module $G$, then uses this representation to produce a set of weights for combining different channels in $E(x)$ to get an attention map per part. Specifically $G(E(x)) \in \R^{M \times C}$, where $M$ is the number of different parts in the image (a set hyperparameter), and the attention map $A_m \in \R^{W \times H}$ for the $m^{th}$ part is computed as

\begin{align}
    A_m(x) = \mathrm{normalize}\left(\mathrm{sigmoid}\left(\sum_{c \in [C]} G_{m,c} \times E_c(x)\right)\right)
\end{align}

where the sum is over the channel dimension, the notation $G_{m,c}$ drops the dependence on $E(x)$ for conciseness and the operation ``normalize'' divides all elements by a constant to make the elements of $A_m$ sum to $1$. The $m^{th}$ part feature $z_m \in \mathbb{R}^C$ is then calculated as:
\begin{align} \label{part-feat}
z_{m,c} = \sum_{w,h} [A_m(x) \odot E_c(x)]_{(w, h)}  , \quad \forall c \in [C]
\end{align}
where $\odot$ is element-wise multiplication and the sum is over the width and height dimensions for each individual channel. We parameterized $E(\cdot)$ using a ResNet-34 backbone (till the $conv5$ block), and $G(\cdot)$ using a fully-connected layer.

As we see from Eq. \ref{part-feat}, $z_m$ can be decomposed into an element-wise product of the attention maps ($A_m(x)$) and the extracted convolutional features ($E(x)$), followed by a sum over the width and height dimensions. For the attention maps to correspond to different parts of an image, we would like them to be compact within a given part and diversely spread over different parts. To encourage this behavior, we use the following two criteria:

\begin{itemize}[leftmargin=8pt,nosep]
    \item \textbf{Compactness.} This criterion encourages the attention map for each part to be concentrated around a peak value and is defined as:
    \begin{align}
        L_{com}({A_m}) &= \sum_{w,h} A_{m}^{w,h}[\|w - w^*\|^2 + \|h - h^*\|^2] \label{eq:loss_Com}
    \end{align}
    where $A_{m}^{w,h}$ is the amplitude of $A_m$ at coordinate $(w,h)$, and $(w^*, h^*)$ is the coordinate of the peak value of $A_m$.
    Note that the loss penalizes high values of $A_m$ that are away from its peak value.
    
    \item \textbf{Diversity.} This criterion encourages the attention maps of the different parts to have peak values at different locations. It is defined as :
    \begin{align}
        L_{div}({A_m}) &= \sum_{w,h} A_{m}^{w,h}[\max_{n: n\ne m}A_n^{w,h} - \zeta] \label{eq:loss_Div}
    \end{align}
    where $\zeta$ is a small margin to ensure training robustness. Note that this loss is high when there are portions of the image where two different attention maps have high values.
\end{itemize}

\noindent Combining the two losses, we get
\begin{align}
    \ell_{part}(x) = \sum_m (L_{com} (A_m (x)) + \lambda_1 L_{div} (A_m (x))) \label{eq:loss_part}
\end{align}
where $\lambda_1$ is a hyperparameter balancing the two criteria.

\subsection{Part-type Likelihood Encoder} \label{subsec:ptle}

We assume an underlying mixture of gaussians distribution which $z_m$ comes from: $$z_m \sim \sum_{k \in [K]} \pi_{k,m} {\cal N}(D_{k,m},\gamma^2 I),$$ where $\pi_{k,m}$ represents the likelihood that the sample belongs to the Gaussian component $k$ with mean $D_{k,m}$ in part $m$.
This means $D_{k,m}$ are prototypical part-types, a convex combination of which using the weights $\pi_{k,m}$ gives the mean of the Gaussian mixture for $z_m$. We also refer to these prototypical part-types $D_{k,m}$ simply as prototypes for part $m$. The above entails an autoencoder implementation, where we can use a projection matrix $P_m \in \R^{K \times C}$ to transform $z_m \in \R^{C}$ to the lower dimensional $\pi_m \in \R^{K}$ (Note that we use a $K \ll C$): 
\begin{align}
    \pi_m(x) = \phi(P_m z_m(x)) \label{eq:pi}
\end{align}
where $\phi(\cdot)$ is a softmax function.

The inverse of this transform $D_{m}^\top \in \R^{K \times C}$ is a matrix containing the prototypes $D_{k,m}$ along its rows and approximately satisfying:
\begin{align}
    z_m(x) \approx D_m^\top \pi_m(x)
    \label{eq:z}
\end{align}
Treating $P_m, D_m \in \R^{K \times C}$ as parameters of an autoencoder, the training objective to learn these low dimensional encodings $\pi_m$ for part feature $z_m$ is as follows:

\begin{align}
    \ell_{ae}(x) = \sum_m ( &\| z_m(x) - D_m^\top \phi(P_m z_m(x)) \|^2 \nonumber \\ 
    &+ \lambda_2 \| P_m \|^2 + \lambda_3 \| D_m \|^2 ).
       \label{eq:prob}
\end{align}
where $\lambda_2$ and $\lambda_3$ are the weights for the regularizers.

\section{Explainability and Explainable Models} \label{sec:explainability}
As mentioned in the introduction explainable models are preferred in many practical scenarios. Before moving on to specific low shot generalization tasks, we elaborate how explainability is a consequence of the RPC encoder and the way it is trained, and even before doing that, we try to define explainability in models in the form of certain properties they should have.

\begin{definition}[Explainable Model]
An Explainable Model is one that has the following three properties: \\
1) \textit{Compact Vocabulary.} Its prediction can be associated with a small number of finite discrete concepts. While, finiteness leaves the actual number of such concepts arbitrary, what we require is motivated by humans posed with choosing among multiple choices. In particular, it is reasonable to assume that for each part we have about half dozen prototypes, and each of these prototypes exhibit about 4-6 variants, and finally there are about 4-6 total number of parts.\\
2) \textit{Causality.} It assigns importance values to each of these concepts such that its final output is directly related to these values; the concepts and importance values together constitute an \emph{explanation}. \\
3) \textit{Meaningfulness: } The concepts must be semantically meaningful to or recognizable by a human either immediately or after seeing a few examples. 
\end{definition}

As per the above properties, our RPC model has a compact vocabulary because it breaks up an image into parts such that each part is representable by only a small number of prototypes/concepts. There is causality, since the model's outputs are a simple function (using the task specific predictor; see Sec \ref{subsec:task-specific}) of the RPC encodings which are simply a set of importance values representing the likelihood that each part comes from a certain concept/prototype. The only component remaining is meaningfulness, which we demonstrate by crowd-sourcing answers to certain questions about example encodings that our model produces for different images (See Sec. \ref{sec:human_eval}).  Using $\ell_{task}$ as a proxy for some loss function related to a classification task (assuming additional possibly parametric functions acting on the RPC encodings to generate a class prediction), when the RPC model gets trained with the loss function $\ell_{part} + \ell_{ae} + \ell_{task}$, it attains the properties mentioned above, making it explainable.

\section{RPC for Low-Shot Generalization} \label{subsec:task-specific}
Using the RPC encoder, a classifier can be defined as a simple multi-layer perceptron (or even a linear classifier) that operates on the RPC encodings $\pi$. The encodings $\pi_m$ from all parts are concatenated before feeding into the classifier.
We specifically used a 2-layer MLP for simply learning a source/training dataset classifier, which we use for evaluating the encodings' interpretability and its adversarial robustness (sections \ref{sec:human_eval} \& \ref{subsec:robustness} respectively). 
Additionally, we evaluated RPC on 3 different low shot generalization tasks, each with its constraints and hence requiring a separate task-specific predictor that we describe next.

\subsection{Unsupervised Domain Adaptation}
This problem involves tackling image classification in a target domain, where the joint distribution of the image-label pairs is different from the training data(\ie $p_{S}(x,y) \neq p_{T}(x,y)$, where $p_S$ is the distribution of the source domain set of labeled training images available to the learner). The problem specifies that $p_S$ and $p_T$ share the same set of labels. An example where this problem arises is when a learner is expected to recognize real images of an object, having only access to labeled hand-sketched depictions of the same. 

Concretely, the learner gets access to a labeled set of images $\D_{S} = \{(x_i^{s}, y_i^{s}) \mid i \in [n_s]\}$, and an unlabeled set of images $\D_{T} = \{x_i^{t} \mid i \in [n_t]\}$, and the goal of the learner is to predict class labels for images in the set $\D_{T}$.

The task-specific predictor in this case, is a simple 2 layer MLP classifier on top of the RPC encodings, $V:\R^{M\cdot K} \rightarrow \R^{|\mc{Y}|}$, where $|\mc{Y}|$ is the number of classes in the set of possible labels $\mc Y$. We first train $V$ along with the rest of the model using only the source domain images to minimize the objective 
\begin{align}
    \E_{(x, y) \sim \D_S} [\ell_{part}(x) + \ell_{ae}(x) + CE(V(\pi(x)), o(y))], \label{eq:loss_DA1}
\end{align}
where $CE(\cdot, \cdot)$ is cross-entropy and $o(y)$, is a one-hot representation of class label $y$. Similar to prior work \cite{chadha2018improving,saito2017asymmetric}, we then pseudo-label the target domain images $x \in \D_T$ with $\tilde{y} = \arg\max_y V(\pi(x))_y$, where $V(\pi(x))_y$ is the $y$-th element in the $V(\pi(x))$ vector and create a set $\widetilde{\D}_T = \{(x, \tilde{y} | x \in \D_T)\}$. We then re-train the entire model to minimize the objective 
\begin{align}
    \E_{(x, y) \sim \D_S \cup \widetilde{\D}_T} [\ell_{part}(x) + \ell_{ae}(x) + CE(V(\pi(x)), o(y))],  \label{eq:loss_DA2}
\end{align}

\subsection{Few Shot Learning}
Few shot learning (FSL) tests a classifier's ability in learning to classify from a few labeled examples. Evaluation typically involves a ``base training set'' $\D_S$ of image-label pairs, that a learner can use for training. At inference time, the learner is provided with ``episodes'' with a small number $\alpha$ of classes and few labeled images (or support images), $\beta$ in number, per class. A certain number of query images are also provided, and the learner is expected to classify them, into one of the $\alpha$ classes, using the support images to possibly adapt itself. This is called an $\alpha$-way $\beta$-shot learning task. The classes in $\D_S$ are typically disjoint from classes used in episodes for evaluation. 

We first train the model with a 2-layer MLP classifier as $V$, on the following loss function
\begin{align}
    \E_{(x, y) \in \D_S} \ell_{part}(x) + \ell_{ae}(x) + CE( V(\pi(x)), o(y))
\end{align}
where $CE$ and $o$ are the same as used in Eq. \ref{eq:loss_DA1}. 

At inference time, we then use the model as a non-parametric nearest neighbors classifier. Let an evaluation episode be notated as follows : $\D_{sup} = \{(x, y) \mid y \in [\alpha]\}$ is the support set of image label pairs, with exactly $\beta$ images per class and $\D_q = \{x_i \mid i \in [q]\}$ is the query set with only images and no labels. For an $x \in \D_q$, the model predicts the label
\begin{align}
    \widehat{y} = \argmin_{y \in [\alpha]} \norm{\pi(x) - \overline{\pi}_y}_2
\end{align}
where $\overline{\pi}_y$ is the average RPC encoding of the class $y$ images in $\D_k$, \ie
\begin{align*}
    \overline{\pi}_y = \frac{\sum_{(x', y') \in \D_{sup}} \mathbbm{1}[y = y']\pi(x')}{\sum_{(x', y') \in \D_{sup}} \mathbbm{1}[y = y']}
\end{align*}

\subsection{Generalized Zero-shot Learning}
Zero-shot learning (ZSL) involves a recognition problem where the class labels for training images have no overlap with the ground truth labels of images seen at inference time. What accompanies this problem is a semantic vector for each class label, that carries meaning for a learner to leverage when it tries to classify images of classes it has not encountered during training. We note here that we will also use the term semantic attributes of a class to refer to this semantic vector, since quite often this vector comes from a labeling of attributes in a dataset.

Specifically, the problem involves a training set of image-label pairs such that $\D_S$ for $(x, y) \in \D_S$, $y \in \mc{Y}^{s} \subset \mc{Y}$. The set $\mc{Y}^{s}$ is often termed ``seen classes'', and the set $\mc{Y}^{u} := \mc{Y} \setminus \mc{Y}^{s}$, ``unseen classes''. As mentioned above, each class label $y \in \mc{Y}$ has a corresponding semantic vector $\sigma_y$. Additionally we notate with $\Sigma^{s}, \Sigma^{u}$ and $\Sigma$, the set of semantic vectors corresponding to $\mc{Y}^{s}, \mc{Y}^{u}$ and $\mc{Y}$. Zero-shot learning refers to the problem where at inference time, images that the model makes predictions on, have classes in $\mc{Y}^{u}$ and the model makes use of the set $\Sigma^{u}$ to make this prediction. A modification of this, called Generalized Zero-shot Learning (GZSL), at inference time, asks the model to predict class labels for images that could be from any class in the set $\mc Y$, using the set of semantic vectors $\Sigma$. ZSL is a simpler problem compared to GZSL since in the former, the learner at inference time, has the knowledge that the correct answer is in the set $\mc Y^{u}$ which is strictly smaller than $\mc Y$ in cardinality. We use GZSL for evaluation. The task predictor $V$ is a 2-layer neural network, and the objective function minimized by the model is
\begin{align}
    \E_{(x, y) \in \D_S} [ &\ell_{part}(x) + \ell_{ae}(x) + \ell_{GZSL}(x, y)]
\end{align}
where
\begin{align}
    \ell_{GZSL}&(x, y) \nonumber \\
    &= \sum_{\substack{y' \in \mc{Y} \\ y' \neq y}} \max(\eta + (\sigma_{y'} - \sigma_{y})^\top V(\pi(x)), 0)
\end{align}
is a hinge loss treating class semantic vectors as the weights of a maximum margin classifier on $V(\pi(x))$.

At inference time, the class prediction for an image $x$ is made as $\widehat{y} = \argmax_{y \in \mc{Y}}\sigma_y^\top V(\pi(x))$.

\subsection{Implementation Details}
Recall from Sec. \ref{subsec:part-feat}, that the feature extractor module $E(\cdot)$ is parametrized by a Resnet-34 (up to the $conv5$ block) and $G(\cdot)$ is a fully connected layer. The number of parts $M$ and the number of prototypes $K$ in each part are hyper-parameters. In our experiments, $M$ is set to 4 and $K$ is set to 16, if not specified. Also, the softmax function $\phi$ in Eq \ref{eq:pi} has a temperature $100$ (selected using the validation set accuracy of few-shot classification on \textit{mini}Imagenet; the sweep showed that accuracy did not change a lot for lower temperatures down to 10, but degraded quickly below a temperature of 1). $\zeta$ in Eq. \ref{eq:loss_Div} is empirically set to a value of 0.02 that achieves robust training. We set $\lambda_2, \lambda_3$ to $1e-3$ in all the experiments. Model optimization for all problems is done in an alternating manner where in step (A) we optimize the parameters of $G(\cdot)$ on the only the objective $\ell_{part}$, and in step (B) we freeze these weights and optimize the rest of the model parameters on the entire (task-specific) objective function.

For FSL and for DA, an input image size of 224 $\times$ 224 pixels is used, and $\lambda_1$ in Eq.(\ref{eq:loss_part}) is set to 2. 
For GZSL, our model takes input image of size 448 $\times$ 448 and $\lambda_1$ is set to 5.
The task-specific predictor $V(\cdot)$ for both GZSL and DA is implemented by a two FC-layer neural network with ReLU activation, the number of neurons in the hidden layer is set to 32. Note that the details of training for each task are mentioned in Sec \ref{sec:experiments}.

\section{Experiments}\label{sec:experiments}

\subsection{Domain Adaptation}
{\noindent \bf Datasets.}
We evaluated the RPC model in unsupervised domain adaptation task between three digits datasets: MNIST\cite{lecun1998gradient}, USPS and SVHN\cite{netzer2011reading}. Each dataset contains 10 classes of digit numbers (0-9). MNIST and USPS are handwritten digits while SVHN is obtained from house number in google street view images.  

{\noindent \bf Setup.}
We follow the same protocol as in \cite{tzeng2017adversarial}, where the three adaptation scenarios used for evaluation are ($\D_S \to \D_T$): MNIST$\to$USPS, USPS$\to$MNIST, and SVHN$\to$MNIST. In the experiments, two variants of our model are evaluated: (1) {\bf RPC(source $\pi$)}: During training, the model is purely learned from source data, which corresponds to the model minimizing the objective in Eq. \ref{eq:loss_DA1}. 
This model does not utilize any information from the unlabeled target data in the training. (2) {\bf RPC(joint $\pi$)}: This model learns the visual encoder from the joint dataset $\mathcal{D}_S\cup\widetilde{\mathcal{D}}_T$ by minimizing the objective in Eq. \ref{eq:loss_DA2}. Recall that $\widetilde{\D}_{T}$ contains images from the target domain with pseudo-labels produced by the ``source $\pi$'' model. %

{\noindent \bf Training Details.}
RPC (source $\pi$) is trained on the source domain dataset, as described above. The learning rate for step (A) and step (B) is 1e-6 and 1e-5. The training epochs are set to be 40, 20, and 40 on MNIST, USPS, and SVHN, respectively. For joint $\pi$, we first initialize with weights with those of the source-$\pi$ model. Next, the model is trained on the joint dataset $\mathcal{D}_S\cup\widetilde{\mathcal{D}}_T$ for 10 epochs. The learning rate for step (B) is modified to 1e-6.

{\noindent \bf Results.}
Target classification accuracies for different scenarios are reported in {\it Table~\ref{tab:da}}. Our RPC approach fares surprisingly well, especially when we use target pseudo-labels to train, \ie, RPC(Joint $\pi$). 
{\color{black}It is worth noting that Mean-Teacher uses a data augmentation technique which models the distortion in target data. Evidently, this technique for the specific dataset is powerful enough that the reported accuracies are higher than those reported for a fully supervised model on target data. In contrast, our method learns a static universal representation for both source and target domain, which does not require the prior knowledge on the domain distortion. The data augmentation is complementary to our model and it can be expected that our model can also benefit from the increased training data.}

\begin{figure}[t]
\centering
\includegraphics[width=\columnwidth]{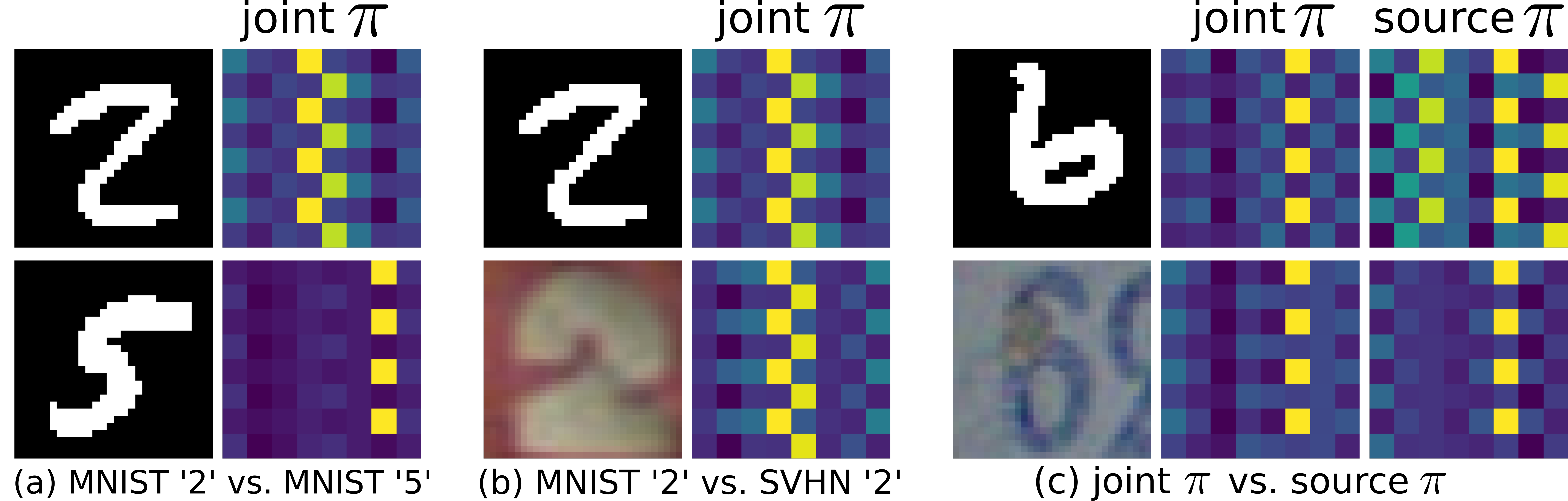}
\caption{Proposed RPC encoding $\pi$ on digit datasets. The $64$-dimensional $\pi$ vector is reshaped to a $8\times8$ matrix for better visualization. For all three examples (a-c), $\pi$ is trained for SVHN$\to$MNIST experiment. }
\label{fig:pi}
\end{figure}

The results demonstrate the benefits of proposed RPC representation. Specifically, in the same domain, the distance or dissimilarity between RPC encodings for different classes are large enough to learn a good classifier. Meanwhile, %
the representations of the same class from different domains are much more similar than the high-dimensional features of a DCN (this can be seen by comparing target accuracies of ``Source only'' and the ``RPC(source $\pi$)'' models), resulting in a similar distribution of features across the two domains. The classifier trained on the source domain is thus able to be used on target domain data. This also means that RPC is more tolerant to visual distortions. 
In Fig. \ref{fig:pi}(a-b), we see that the encodings learnt by RPC(joint $\pi$) model are quite similar for images of the same class across the two domains.

{\noindent \it Source vs. Joint $\pi$}.
Comparing accuracies in Table \ref{tab:da}, we see that RPC(joint $\pi$) has better performance, showing that cross-entropy loss using pseudo-labels on the target domain in Eq. \ref{eq:loss_DA2} helps. This model benefits more over RPC(source $\pi$) when the domain shift is severe (\eg S $\rightarrow$ M). This is also clear from a visual representation of the encodings in Figure \ref{fig:pi}(c), where we see a larger difference in encodings across the two domains using RPC(source $\pi$) than RPC(joint $\pi$).

\begin{table}[t]
    \centering
    \renewcommand{\arraystretch}{1.16}
    \setlength{\tabcolsep}{0.15cm}
    \begin{tabular}{l c c c}
        \toprule
        \bf Methods & M $\rightarrow$ U & U $\rightarrow$ M & S $\rightarrow$ M \\
        \midrule
        Source Only$^{*}$ & 75.2 & 57.1 & 60.1 \\
        Gradient reversal$^{*}$~\cite{ganin2016domain} & 77.1 & 73.0 & 73.9\\
        Domain confusion$^{*}$~\cite{tzeng2015simultaneous} & 79.1 & 66.5 & 68.1 \\ 
        CoGAN$^{*}$~\cite{liu2016coupled} & 91.2 & 89.1 & - \\
        ADDA$^{*}$~\cite{tzeng2017adversarial}  & 89.4 & 90.1 & 76.0 \\
        DTN~\cite{taigman2016unsupervised} & - & - & 84.4 \\
        UNIT~\cite{liu2017unsupervised} & 96.0 & 93.6 &  90.5 \\
        CyCADA~\cite{hoffman2017cycada} & 95.6 & 96.5 & 90.4 \\
        MSTN~\cite{xie2018learning} & 92.9 & - & 91.7 \\
        ADR~\cite{saito2017adversarial} & 91.3 & 91.5 & 94.1 \\
        MCD~\cite{saito2018maximum} & 94.2 & 94.1 & 96.2 \\
        CDAN~\cite{long2018conditional} & 95.6 & \color{blue}{\bf{98.0}} & 89.2 \\
        Mean-Teacher~\cite{french2017self} & \color{blue}{\bf 98.3} & \bf{\color{red}99.5} & \bf{\color{red}99.3}\\
        SHOT~\cite{liang2020we} & 97.9 & \color{blue}{\bf{98.0}} & \bf{\color{blue}98.9} \\
        \midrule
        RPC (source $\pi$) & 94.8 &  96.1 & 82.4\\
        RPC (joint $\pi$) & \color{red}{\bf{98.8}} & 96.8 & 95.2 \\
        \bottomrule
    \end{tabular}
\caption[caption]{Domain adaptation classification results. M = MNIST, U = USPS, S = SVHN. The highest accuracy is in \textcolor{red}{\bf red} color and the second is in \textcolor{blue}{\bf blue} (better viewed in color). Self-ensembling, unlike other methods, leverages data-augmentation and reports accuracy numbers that are evidently higher than those obtained in the fully supervised case for $U\rightarrow M,\,S \rightarrow M$. $^{*}$Numbers reported in \cite{tzeng2017adversarial}. \label{tab:da}}
\end{table}

\subsection{Few-Shot Learning}
{\noindent \bf Datasets.} We first evaluate the few shot learning performance of the proposed model on three benchmark datasets: Omniglot\cite{lake2015human}, {\it mini}ImageNet\cite{vinyals2016matching} and CUB~\cite{WahCUB_200_2011}. Omniglot consists of 1623 characters from 50 alphabets. Each character (class) contains 20 handwritten images from people. {\it mini}ImageNet is a subset of ImageNet\cite{ILSVRCarxiv14} which contains 60,000 images from 100 categories. The CUB dataset contains 200 classes each corresponding to a different species of bird and a total of 11,788 images.

{\noindent \bf Setup.} We follow the same protocol as \cite{sung2018learning} for the first two benchmarks. For Omniglot, the dataset is augmented with new classes through 90\degree, 180\degree\ and 270\degree rotations of existing characters. 1200 original classes plus rotations are selected as training set and the remaining 423 classes with rotations are test set. For {\it mini}ImageNet, the dataset is split into 64 training, 16 validation and 20 testing classes. For the CUB dataset, we followed \cite{chen2019closer} and used 100 classes for training, 50 for validation and 50 for testing. The model is only trained on the training set and the validation set is used for development.

We evaluate the 5-way accuracy on {\it mini}ImageNet and 5-way plus 20-way accuracy on Omniglot. 1-shot and 5-shot learning performance is evaluated in each setting. For {\it $\alpha$-way $\beta$-shot} learning, in each test episode, $\alpha$ classes will be randomly selected from the test set, then {\it k} samples will be drawn from these classes as support examples, and 15 examples will be drawn from the rest images to construct the test set. We run 1000 and 600 test episodes on Omniglot and {\it mini}ImageNet, respectively, to compute the average classification accuracy.

{\noindent \bf Training Details.} Our model is trained for 80, 10 and 100 epochs on Omniglot, {\it mini}ImageNet and CUB, respectively. The learning rate for step (A) is set to 1e-6, and the learning rate of step (B) is 1e-4 for Omniglot. The learning rate is set to 1e-4 for {\it mini}ImageNet and 5e-4 for CUB. On {\it mini}ImageNet and CUB, we use $K=256$ prototypes for few-shot classification. For few-shot classification on {\it mini}ImageNet, we used a Resnet-18 feature extractor as opposed to the Resnet-34 used for other experiments, and the weights for the feature extractor 
$E(\cdot)$ are pre-trained on an appropriate Imagenet subset following~\cite{bateni2020improved}, while they are randomly initialized for Omniglot and CUB.

\begin{table}[t]
    \centering
    \small
    \renewcommand{\arraystretch}{1.3}
    \setlength{\tabcolsep}{0.1cm}
    \begin{tabular}{l c c c c}
        \toprule
        \multirow{2}{*}{\bf Methods}
        & \multicolumn{2}{c}{\textbf{5-way Acc.}} & \multicolumn{2}{c}{\textbf{20-way Acc.}} \\
        & \it 1-shot & \it 5-shot & \it 1-shot & \it 5-shot \\
        \hline
        Matching Net~\cite{vinyals2016matching} & 98.1 & 98.9 & 93.8 & 98.5 \\
        MAML~\cite{finn2017model} & 98.7 & \bf\color{red} 99.9 & 95.8 & 98.9 \\
        Prototypical Net~\cite{snell2017prototypical} & 98.8 & 99.7 & 96.0 & 98.9 \\
        Relation Net~\cite{sung2018learning} & 99.6 & \bf\color{blue}99.8 & 97.6 & 99.1 \\
        GCR~\cite{li2019few} & \bf\color{blue}99.7 & \bf\color{red}99.9 & \bf\color{red}99.6 & 99.3 \\
        TapNet~\cite{yoon2019tapnet} & - & - & 98.1 & \bf\color{blue}99.5 \\
        DCN6-E~\cite{liu2021decoder} & \bf\color{red}99.9 & \bf\color{red}99.9 & \bf\color{blue}99.1 & \bf\color{red}99.6 \\
        \midrule
        RPC & 98.9 & \bf\color{blue}99.8 & 96.5 & 99.3 \\
        \bottomrule
    \end{tabular}
    \caption{Few-shot classification accuracy on Omniglot. In each column, \textcolor{red}{\bf red}=highest and \textcolor{blue}{\bf blue}=2$^{nd}$ highest accuracy \label{tab:fsl-omniglot}}
\end{table}

\begin{table}[]
    \centering
    \renewcommand{\arraystretch}{1.1}
    \setlength{\tabcolsep}{1mm}
    \begin{tabular}{l c c}
    \toprule
    \bf Methods & \it 1-shot & \it 5-shot \\
    \midrule
        Baseline++~\cite{chen2019closer} & 51.87 & 75.68 \\
        ProtoNet~\cite{snell2017prototypical} & 54.16 & 73.68 \\ 
        MetaOptNet~\cite{lee2019meta} & 64.09 & 80.00 \\
        New-Meta~\cite{chen2020new} & 63.17 & 79.26 \\
        FEAT~\cite{ye2020few} & 66.78 & 82.05 \\
        DeepEMD~\cite{zhang2020deepemd} & 65.91  & 82.41 \\
        DPGN~\cite{yang2020dpgn} & 67.77 & 84.60 \\
        RENet~\cite{kang2021relational} & 67.60 & 82.58 \\
        ZN~\cite{fei2021z} & 67.35 & 83.04 \\
        MeTAL~\cite{baik2021meta} & 66.61 & 81.43 \\
        BML~\cite{zhou2021binocular} & 67.04 & 83.63 \\
        ECSIER~\cite{rizve2021exploring} & 67.28 & 84.78 \\
        ECKPN~\cite{chen2021eckpn} & \bf\color{blue}70.48 & \bf\color{blue}85.42 \\
        Simple-CNAPS~\cite{bateni2020improved} & \bf \color{red} 77.40  & \bf\color{red}90.30 \\
    \midrule
    RPC & 63.92 & 84.57 \\
    \bottomrule
    \end{tabular}
    \caption{Few-shot accuracy in \% on \textit{mini}ImageNet. In each column, \textcolor{red}{\bf red}=highest and \textcolor{blue}{\bf blue}=2$^{nd}$ highest accuracy 
    }
    \label{tab:fsl-miniin}
\end{table}

\begin{table}[tbp!]
    \centering
    \renewcommand{\arraystretch}{1.1}
    \setlength{\tabcolsep}{2mm}
    \begin{tabular}{l c c c}
    \toprule
    \bf Methods & \it 1-shot & \it 5-shot \\
    \midrule
    Baseline++~\cite{chen2019closer} & 68.00 & 84.50 \\
    ProtoNet~\cite{snell2017prototypical} & 72.94 & 87.86 \\
    SimpleShot~\cite{wang2019simpleshot} & 68.90 & 84.01 \\
    DN4$\dagger$~\cite{li2019revisiting} & 70.47 &84.43\\
    COMET~\cite{cao2020concept} & 72.20 & 87.60 \\
    MetaOptNet*~\cite{lee2019meta} & 75.15 & 87.09 \\
    DeepEMD~\cite{zhang2020deepemd} & 75.65 & 88.69 \\
    AFHN\cite{li2020adversarial} & 70.53 & 83.95 \\
    BSNet\cite{li2020bsnet} & 69.61 & 83.24 \\
    MTL*\cite{liu2018meta} & 73.31 & 82.29 \\
    VFD* \cite{xu2021variational} & 79.12 & \bf\color{blue}91.48 \\
    FOT\cite{wang2021fine} & 72.56 & 87.22 \\
    FRN\cite{wertheimer2021few} & \bf\color{red} 83.16 &  \bf\color{red}92.59 \\ %
    RENet\cite{kang2021relational} & \bf\color{blue} 79.49 & 91.11 \\
    \midrule
    RPC & 75.01 & 89.61 \\
    \bottomrule
    \end{tabular}
    \caption{Few-shot accuracy in \% on CUB. If not specified, the results reported by the original paper. *: reported in \cite{xu2021variational}. $\dagger$: results are obtained using the authors' implementation. In each column, \textcolor{red}{\bf red}=highest and \textcolor{blue}{\bf blue}=2$^{nd}$ highest accuracy}
    \label{tab:fsl-cub}
\end{table}

\begin{table*}[t]
    \centering
\renewcommand{\arraystretch}{1}
\setlength{\tabcolsep}{0.3cm}
\scalebox{1.0}{
    \begin{tabular}{l| c c c| c c c| c c c }
        \toprule
        \multirow{2}{*}{\bf{Methods}} & \multicolumn{3}{c|}{\bf{CUB}} & \multicolumn{3}{c|}{\bf{AWA2}} & \multicolumn{3}{c}{\bf{aPY}} \\
         & U & S & H & U & S & H & U & S & H\\
        \midrule
        SJE\cite{akata2015evaluation} & 23.5 & 59.2 & 33.6 & 8.0 & 73.9 & 14.4 & 3.7 & 55.7 & 6.9 \\
        SAE\cite{kodirov2017semantic} & 7.8 & 54.0 & 13.6 & 1.1 & 82.2 & 2.2 & 0.4 &   \bf{\color{red}80.9} & 0.9 \\
        SSE\cite{zhang2015zero} & 8.5 & 46.9 & 14.4 & 8.1 & 82.5 & 14.8 & 0.2 & 78.9 & 0.4 \\
        ALE\cite{akata2016label} & 23.7 & 62.8 & 34.4 & 14.0 & 81.8 & 23.9 & 4.6 & 73.7 & 8.7 \\
        SYNC\cite{changpinyo2016synthesized} & 11.5 & 70.9 & 19.8 & 10.0 & 90.5 & 18.0 & 7.4 & 66.3 & 13.3 \\
        PSRZSL\cite{Annadani_2018_CVPR} & 24.6 & 54.3 & 33.9 & 20.7 & 73.8 & 32.3 & 13.5 & 51.4 & 21.4 \\
        SP-AEN\cite{chen2018zero} & 34.7 & 70.6 & 46.6 & 23.3 & 90.9 & 37.1 & 13.7 & 63.4 & 22.6 \\
        CE-GZSL\cite{han2021contrastive} & 63.9 & 66.8 & 65.3 & 63.1 & 78.6 & 70.0 & - & - & - \\
        GEM-ZSL\cite{liu2021goal} & \bf \color{blue}64.8 & \bf \color{blue}77.1 & {\bf \color{blue}70.4} & 64.8 & 77.5 & 70.6 & - & - & - \\
        \midrule
        {\it Generative ZSL} & & & & & & & & & \\
        GDAN\cite{huang2018generative} & 39.3 & 66.7 & 49.5 & 32.1 & 67.5 & 43.5 & \bf {\color{blue}30.4} & 75.0 & \bf {\color{blue}43.4} \\
        CADA-VAE\cite{schonfeld2018generalized} & 51.6 & 53.5 & 52.4 & 55.8 & 75.0 & 63.9 & - & - & - \\
        LisGAN \cite{li2019leveraging} & 46.5 & 57.9 & 51.6 & - & - & - & 34.3 & 68.2 & 45.7 \\
        f-CLSWGAN \cite{xian2018feature} & 43.7 & 57.7 & 49.7 & - & - & - & - & - & - \\
        SE-GZSL\cite{Verma_2018_CVPR} & 41.5 & 53.3 & 46.7 & \bf \color{blue}{58.3} & 68.1 & 62.8 & - & - & - \\
        DA-GZSL\cite{atzmon2018domain} & 47.9 & 56.9 & 51.8 & - & - & - & - & - & - \\
        \midrule
        {\it Trans-ZSL} & & & & & & & & & \\
        DIPL\cite{NIPS2018_7380} & 41.7 & 44.8 & 43.2 & - & - & - & - & - & - \\
        TEDE\cite{zhang2018effective} & \bf \color{blue}{54.0} & 62.9 & 58.1 & \bf\color{blue}{68.4} & \bf\color{red}{93.2} & \bf\color{blue}{78.9} & 29.8 & \bf \color{blue} 79.4 & 43.3 \\
        STHS\cite{bo2021hardness} & \bf \color{red}77.4 & 74.5 & \bf \color{red} 75.9 & \bf \color{red} 94.9 & \bf \color{blue} 92.3 & \bf \color{red} 93.6 & - & - & - \\
        \midrule
        RPC & 33.4 & {\bf\color{red}87.5} & 48.4 & 41.6 & 91.3 & 57.2 & 24.5 & 72.0 & 36.6 \\
        RPC + CS & 59.2 & 74.6 & 66.0 & 54.6 & 87.7 & 67.3 & {\bf\color{red}41.1} & 68.0 & {\bf\color{red}51.2} \\
        \bottomrule
    \end{tabular}}
    \caption{GZSL results on CUB, AWA2 and aPY. U = unseen classes, S = seen classes, H = harmonic mean. The accuracy is class-average Top-1 in \%. In each column, \textcolor{red}{\bf red}=highest and \textcolor{blue}{\bf blue}=2$^{nd}$ highest accuracy}
    \label{tab:gzsl}
\end{table*}

{\noindent \bf Results.} Few shot learning results for the different benchmarks are reported in  \textit{Tables}~\ref{tab:fsl-omniglot}, \ref{tab:fsl-miniin} and \ref{tab:fsl-cub}. Again, our RPC approach, which simply trained a classifier on base classes and used the learned encoder, along with nearest neighbors classification, is competitive on the benchmark and does not lag by much compared to the most recent state of the art methods on this task. We interpret this as a result of the low intra class and the high inter class divergences that are inherently exhibited by the encodings generated by our RPC trained classifier, allowing for the use of nearest neighbors for few-shot inference. The high dimensional feature spaces of DCNs do not exhibit these nice properties, and need specific training strategies to function well in this problem.

\subsection{Generalized Zero-Shot Learning}

{\noindent \bf Datasets.}
The performance of our model for GZSL is evaluated on three commonly used benchmark datasets: {\it Caltech-UCSD Birds-200-2011} (CUB) \cite{WahCUB_200_2011}, {\it Animals with Attributes 2} (AWA2) \cite{xian2018zero} and Attribute Pascal and Yahoo (aPY) \cite{farhadi2009describing}. CUB is a fine-grained dataset consisting of 11,788 images from 200 different types of birds. 312-dimensional semantic attributes are annotated for each category. AWA2 has 37,222 images from 50 different animals and 85-dim class-level semantic attributes. aPY contains 20 Pascal classes and 12 Yahoo classes. It has 15,339 images in total and 64-dimensional semantic attributes are provided. We did not choose another popular GZSL benchmark dataset, SUN \cite{xiao2010sun}, for the reason that the scene images in SUN are not typical objects that can be decomposed into our part-prototype hierarchy.

{\noindent \bf Setup.}
It has been shown\cite{xian2018zero} that the conventional ZSL setting is overly optimistic because it leverages absence of seen classes at test-time and there is consensus that methods should focus on the generalized ZSL setting. We thus, use GZSL for our evaluations.
Following the protocol in \cite{xian2018zero}, we evaluated the average-class Top-1 accuracy on unseen classes (U), seen classes (S) and the harmonic mean (H) of S and U.

{\color{black} It has been observed that in GZSL, a classifier trained using seen class images often predicts output class probabilities that are higher for seen classes than unseen \cite{Chao2016AnES}, which results in poor performance. Calibrated Stacking(CS) is proposed in \cite{Chao2016AnES} to balance the performance between seen and unseen classes by calibrating the scores of seen classes. Hence, in addition to our original model, we also evaluated with CS (denoted as RPC+CS in Table \ref{tab:gzsl}). The parameters for CS were chosen using cross validation.}

{\noindent \bf Training Details.}
Our models are trained for 120, 100 and 110 epochs on CUB, AWA2 and aPY, respectively. The learning rate for step (A) is set to 1e-6 and that for step (B) is set to 1e-5.

{\noindent \bf Methods for Comparison.} 
We list here, GZSL methods that we compared RPC with in Table \ref{tab:gzsl}. Comparisons are not all apples-to-apples since some of these approaches use different assumptions: {\color{black} (1) Methods in the top section, learn a compatibility function between the visual and semantic representations: SJE\cite{akata2015evaluation}, ALE\cite{akata2016label}, SAE\cite{kodirov2017semantic}, SSE\cite{zhang2015zero}, SYNC\cite{changpinyo2016synthesized}, PSRZSL\cite{Annadani_2018_CVPR}, SP-AEN\cite{chen2018zero}, CE-GZSL\cite{han2021contrastive}, and GEM-ZSL\cite{liu2021goal}. {\bf Our method also uses this strategy.} (2) Generative model based methods ({\it Generative-ZSL}) synthesize unseen examples or features using generative models like GAN and VAE thus requiring unseen class semantics at training time: GDAN\cite{huang2018generative}, CADA-VAE\cite{schonfeld2018generalized}, 3ME\cite{felix2019multi}, SE-GZSL\cite{Verma_2018_CVPR}, LisGAN\cite{li2019leveraging}, f-CLSWGAN\cite{xian2018feature}, and DA-GZSL\cite{atzmon2018domain}.
(3) Transductive ZSL methods ({\it Trans-ZSL}) methods work in a transductive setting which allows the model access to unlabeled images from unseen classes during training: DIPL\cite{NIPS2018_7380}, TEDE\cite{zhang2018effective}, and STHS\cite{bo2021hardness}.}

{\noindent \bf Results.}
{\color{black} Results for GZSL are in Table~\ref{tab:gzsl}. Without calibrated stacking, comparing the harmonic mean (H) accuracy, we see RPC outperforms all other compatibility function based methods (the first section of the table). After the scores are calibrated, our model (RPC+CS) obtains 66.0\%, 67.3\% and 51.2\% for the harmonic mean, respectively, which outperforms other approaches compared to, except TEDE on AWA2. 

It is worth noting that, the {\it Generative ZSL} and {\it Trans-ZSL} methods always obtain higher accuracy than compatibility function methods, except for our models. This is because the generative and trans-ZSL methods have access to additional information of unseen classes during training. However, this assumption is too optimistic in a real world ZSL scenario since it is unlikely to have full knowledge of all unseen categories at training time. In contrast, our models can be applied in the scenario where novel classes may only appear at test time. Still, by only leveraging seen classes knowledge, our RPC model obtains competitive and sometimes even better performance than generative and trans-ZSL methods.}

\begin{figure}
\centering
\includegraphics[width=\columnwidth]{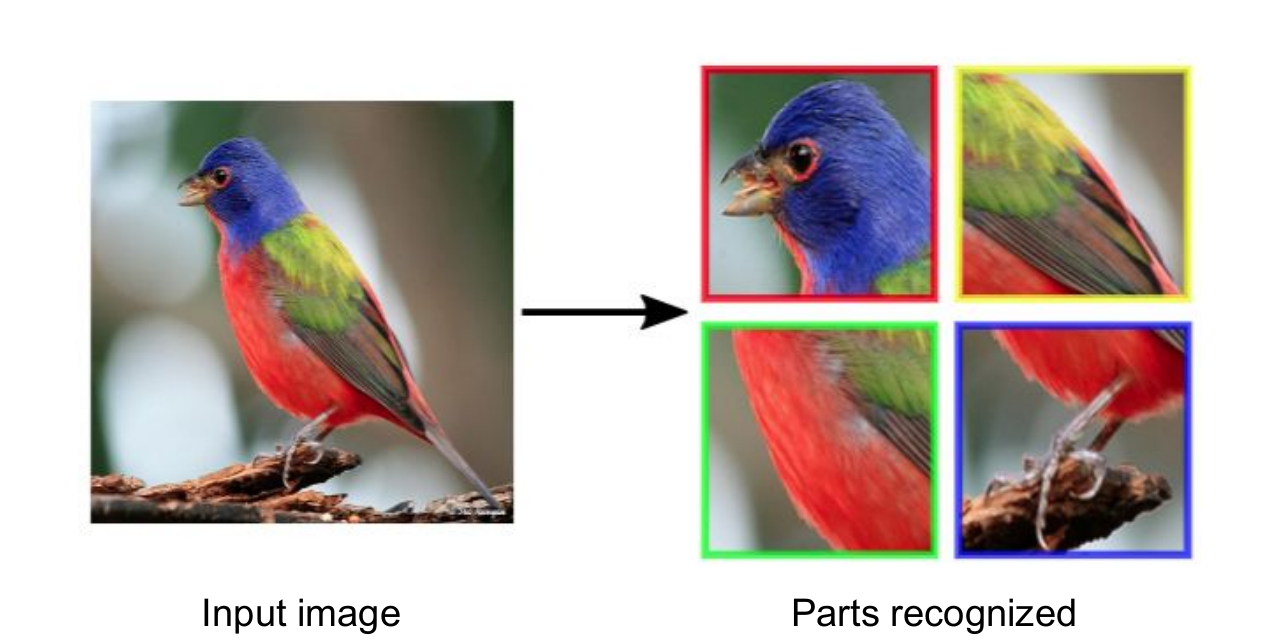}
\caption{Example parts recognized by our model in the image of a ``Painted Bunting'' from the CUB dataset. These parts closely correspond to common semantic attributes labeled in the dataset like ``crown color'', ``feet shape'' etc.}
\label{fig:rpc_bird_parts}
\end{figure}

The success of our model can be attributed primarily to the proposed RPC encodings, which close the gap between the images and their semantic attributes. For example, in Figure~\ref{fig:rpc_bird_parts}, we visualize the part attentions discovered by our model and several semantic attributes for the class {\it `Painted\_Bunting'} in CUB dataset. Our model learns the part areas around ``head'', ``wing'', ``body'', and ``feet'', which correspond to most semantic attribute annotations in the dataset (e.g. crown color:blue, wing color:green, etc.). Using the RPC encodings, our visual attributes mirror the representation of semantic vectors, thus mitigating the large gap between the semantic attributes and high-dimensional visual features learned by DCNs.

\subsection{Robustness to Adversarial Attack} \label{subsec:robustness}
As mentioned in Sec. \ref{sec:related_work}, Szegedy~\etal~\cite{szegedy2013intriguing}, first discovered the susceptibility of deep neural networks to such attacks. In this section we show that our RPC encodings are inherently more robust to adversarial attacks than a deep convolutional classifier trained for the same task. We demonstrate this by choosing a simple FGSM attack \cite{goodfellow2014explaining} (as described next) on models trained for the classification task. No techniques were used in the compared models to specifically train them to be adversarially robust. Note that a range of different attacks and defences have been developed since \cite{szegedy2013intriguing}, but a simple FGSM attack, serves the purpose of demonstrating that our RPC model learns encodings which are less susceptible to small adversarial image distortions.

{\noindent \bf Datasets.} We evaluated on two fine-grained classification datasets, CUB \cite{WahCUB_200_2011} and Stanford Car \cite{KrauseStarkDengFei-Fei_3DRR2013}. The Car dataset contains 16,185 images from 196 different car models. The accuracies on the official test splits are reported.

\begin{figure*}[h]
    \centering
     \begin{subfigure}[b]{0.45\linewidth}
         \centering
         \includegraphics[width=\linewidth]{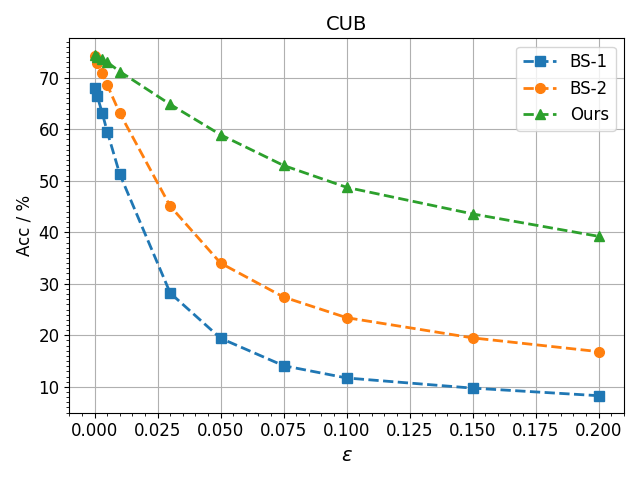}
         \caption{CUB}
         \label{fig:adversarial_cub}
     \end{subfigure}
     \hfill
     \centering
     \begin{subfigure}[b]{0.45\linewidth}
         \centering
        \includegraphics[width=\linewidth]{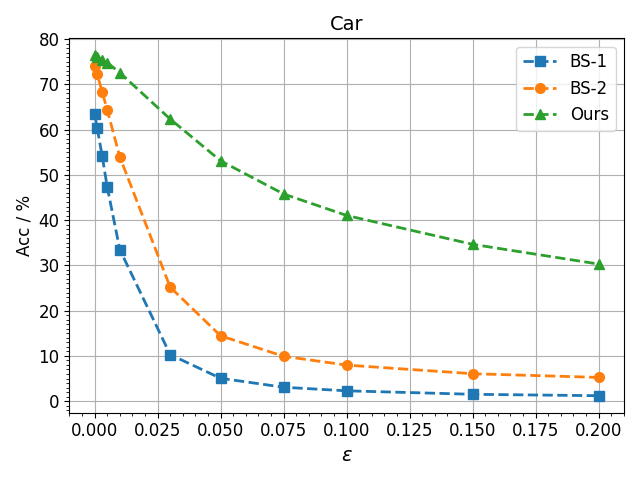}
        \caption{Car}
        \label{fig:adversarial_car}
     \end{subfigure}
     \hfill
     \begin{subfigure}[b]{0.95\linewidth}
         \centering
         \includegraphics[width=\linewidth]{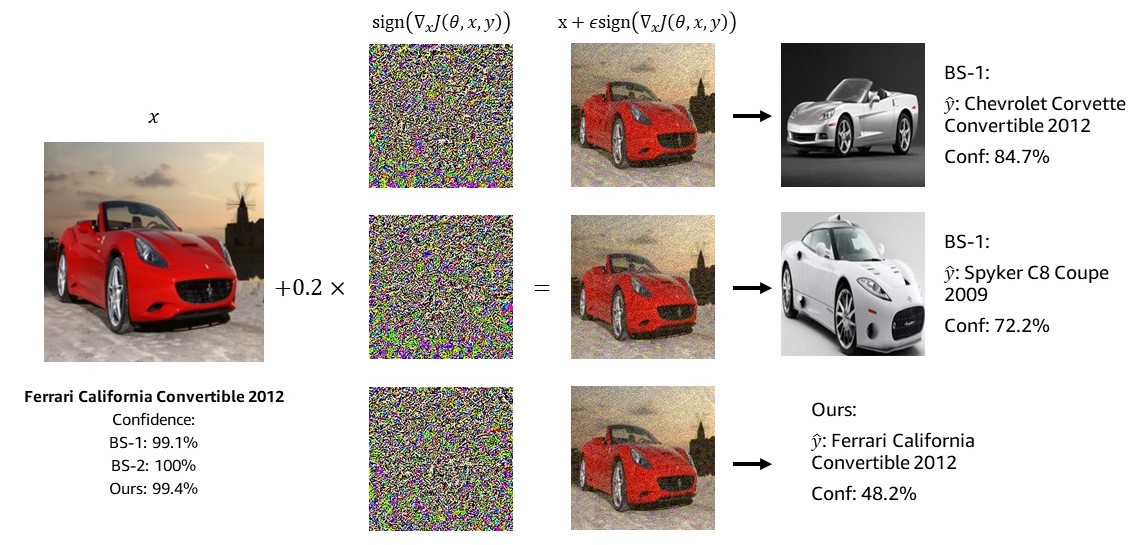}
         \caption{Example of perturbed images from Car dataset. The original  (``Ferrari California Convertible 2012'') were recognized by all three models with $>$99\% confidence. The distortion $\eta$ (Eq.~\ref{eq:fgsa}) to the input is imperceivable to human eys. The two baseline models recognize the perturbed input as incorrect classes ``Chevrolet Corvette Convertible 2012'' and ``Spyker C8 Coupe 2009'', while our model predict the correct class label. RPC achieves this by virtue of encoding images in a manner similar to how humans would.} 
         \label{fig:adv_demo_car}
     \end{subfigure}
     \caption{Robustness to adversarial attacks. (a-b): Test accuracy under various level of FGSA perturbations on CUB (a) and Car (b). (c): Example perturbed images and predictions for baselines and our models. $\epsilon$: the perturbation intensity.}
     \label{fig:adversarial}
\end{figure*}

{\noindent \bf Adversarial Attack.} We use Fast Gradient Sign Attack (FGSA) \cite{goodfellow2014explaining} as the attacker. FGSA is a white-box attacker that has full knowledge and access to the model. Albeit simple, it is a powerful adversarial attack that affects a wide range of classification models. For an image $x$, FGSA generates adversarial perturbation $\mathbf{\eta}$ by calculating the gradient over the input $x$ w.r.t. the cost $J(\theta, x, y)$ used to train the model:
\begin{equation}
    \mathbf{\eta} = \epsilon \mathrm{sign}(\nabla_x J(\theta, x, y)) \label{eq:fgsa}
\end{equation}
where $\theta$ is the parameters of the model, and $\epsilon$ controls the intensity of the perturbations. The perturbed image $x + \eta$ often looks visually similar to $x$ to the human eye, but different models misclassify $x + \eta$, even when their prediction on $x$ is correct (see Fig~\ref{fig:adv_demo_car}).

In our experiments $J(\cdot)$ is implemented by cross-entropy loss for all the models. We evaluate the test accuracy under different distortion level $\epsilon$.

{\noindent \bf Models Compared.} We compare the robustness of RPC with two baseline models:
\begin{itemize}[leftmargin=8pt,nosep]
    \item BS-1 is an adapted ResNet-34 model that has a two-layer MLP as the classifier. It does not have the multi-attention module $G$ and the part-type likelihood encoder compared to our model. BS-1 is trained using a standard cross-entropy loss.
    \item BS-2 has the same MACNN architecture as our model, but the part-type likelihood encoder is absent. The part features $z_m$ for different parts are concatenated and input to the classifier (a 2-layer MLP). BS-2 is trained using the loss $\ell_{part}$ (Eq. \ref{eq:loss_part}) and the cross-entropy loss, using the same alternating optimization strategy as our model.
\end{itemize}

{\noindent \bf Training Details.} BS-1 is trained using a learning rate 5e-5 for 100 epochs on CUB and 110 epochs on Car. The learning rate is decayed by a factor of 0.5 at epochs [60, 80] for CUB and epochs [70, 90] for Car. BS-2 is trained with learning rate 2e-4 on CUB and 4e-4 on Car. It is trained for 60 epochs and the learning rate is decayed by 0.5 every 20 steps. Our RPC model initialized its parameters of the part feature extractor from the trained BS-2 weights. It then is only trained for 5 more epochs with a learning rate 2e-5.

{\noindent \bf Results.} We sweep $\epsilon$ (in Eq \ref{eq:fgsa}) from 0 to 0.2 and evaluated the test accuracy for the three models. The results are plotted in Fig.~\ref{fig:adversarial}(a-b). We see that performs decays at a much lower rate with attacks of increasing intensity for our RPC model as compared to the baselines.
Specifically, when $\epsilon = 0.2$, on CUB, BS-1 has less than 10\% accuracy and BS-2 is less than 20\%, while our model can still achieve ~40\%. On the Car dataset, BS-2 only obtain ~5\% and BS-1 even drops to ~1\%, but our RPC model can maintain an impressive 30\% accuracy. In Fig.~\ref{fig:adv_demo_car} we illustrate an example that recognized by our model is misclassified by the two baseline models with the same distortion intensity. Notice that BS-2 is still very vulnerable to the adversarial attack, even though it has the same multi-attention mechanism in RPC. Because RPC learns to represents each image part in the vocabulary of a small number of prototypes, it is less sensitive to the perturbations in the input image compared to other high-dimensional visual features.

\subsection{Ablating Training Objectives}

For training our RPC encoder, we used two penalties $L_{com}$ and $L_{div}$ that correspond to the compactness and diversity priors on part locations as mentioned in section \ref{sec:method}. Here, we demonstrate the effect that each of these losses has on the model performance and the encodings themselves.

Additionally, the features $z_m(x)$ are encoded into part-type likelihood scores of a gaussian mixture model constituting the RPC code $\pi(x)$ (see Eq. \ref{eq:pi}). The autoencoder loss from Eq. \ref{eq:prob} was used for training the parameters in this mapping. We also demonstrate the effect of removing $L_{ae}$ and using the part features $z_m(x)$ (instead of $\pi_m(x)$) for the classifier.

\begin{table}[h]
    \centering
    \small
    \renewcommand{\arraystretch}{1.3}
    \setlength{\tabcolsep}{0.1cm}
    \begin{tabular}{c c c | c c | c | c}
        \toprule
        \multirow{2}{*}{$L_{com}$} & \multirow{2}{*}{$L_{div}$} & \multirow{2}{*}{$L_{ae}$} & \multicolumn{2}{c|}{FSL CUB} & DA & GZSL CUB \\
        & & & \it 1-shot & \it 5-shot & S$\rightarrow$M & Harmonic mean \\
        \midrule
        \textcolor{dartmouthgreen}{\faCheckSquare} & $\square$ & \textcolor{dartmouthgreen}{\faCheckSquare} & 74.72 & 88.83 & 94.2 & 43.7 \\
        $\square$ & \textcolor{dartmouthgreen}{\faCheckSquare} & \textcolor{dartmouthgreen}{\faCheckSquare} & 73.01 & 87.74 & 94.5 & 39.6 \\
        \textcolor{dartmouthgreen}{\faCheckSquare} & \textcolor{dartmouthgreen}{\faCheckSquare} & $\square$ & 71.84 & 85.63 & 94.4 & 48.1 \\
        \midrule
        \textcolor{dartmouthgreen}{\faCheckSquare} & \textcolor{dartmouthgreen}{\faCheckSquare} & \textcolor{dartmouthgreen}{\faCheckSquare} & 75.01 & 89.61 & 95.2 & 48.4 \\
        \bottomrule
    \end{tabular}
    \caption{Performance metrics on different benchmarks (5-way few shot classification on CUB, domain adaptive classification on transfer from SVHN $\rightarrow$ MNIST, and the harmonic mean accuracy on seen and unseen classes in the case of GZSL on the CUB dataset) when different losses in RPC training are ablated. \label{tab:ablations}}
\end{table}

\begin{figure}[h]
    \centering
    \includegraphics[width=0.95\linewidth]{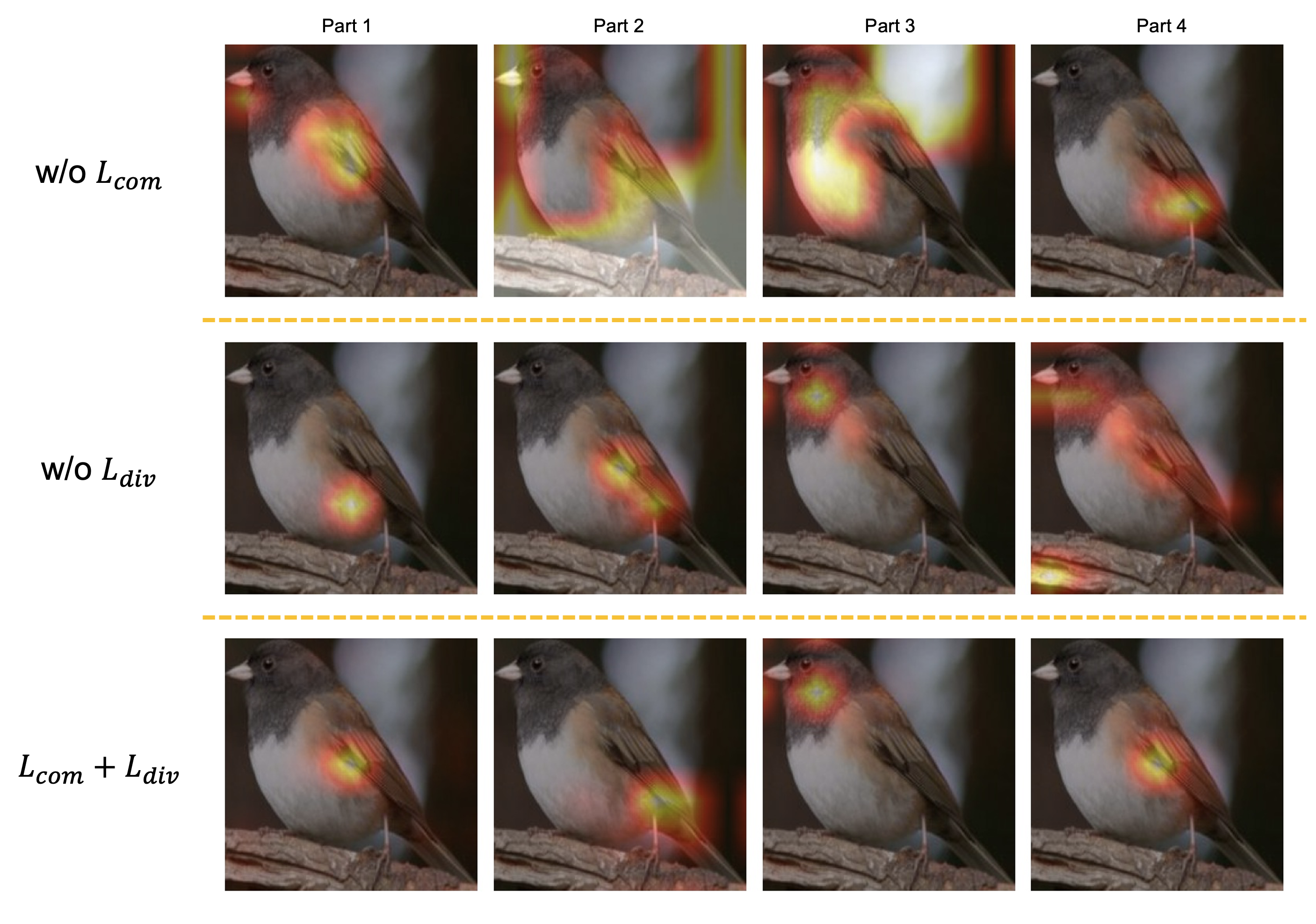}
    \caption{Attention maps of models learned with different objectives from CUB. Row 1: $L_{com}$ is not used; Row 2: $L_{div}$ is not used; Row 3: both $L_{com}$ and $L_{div}$ are used.}
    \label{fig:vis_loss_interpretability}
\end{figure}

In Table \ref{tab:ablations} we can see that the three major components of RPC $L_{ae}$, $L_{com}$ and $L_{div}$ play a positive role when it comes to performance in low shot generalization tasks.  In Fig. \ref{fig:vis_loss_interpretability} %
we see that the two losses on part attentions have intended effects and that when we remove $L_{com}$, the part attention maps are more dispersed, than concentrated at the part locations. When we remove $L_{div}$, the part locations detected by the model can end up on top of each other since each part would seek just the most salient feature disregarding where other parts lie. 

\noindent {\it Robustness.} In passing we also observe from Sec.~\ref{subsec:robustness} that our scheme with the auto-encoder, diversity and compactness also plays a significant role in improving robustness relative to the baseline schemes that do not have the auto-encoder in particular. The auto-encoder, together with compactness and diversity ensure, that the RPC encoder allows for discretization of concepts, and thus for an attack to be successful, the adversary must modify the input image significantly to modify a concept.

\section{Human Evaluation} \label{sec:human_eval}
In this section, we present results of crowd-sourced experiments conducted using Amazon Mechanical Turk (MTurk), that indicate that our RPC encodings are interpretable and agreeable with human perception. We designed three questions to gauge this agreement along different aspects : 1) {\it Discriminability of parts}: Are the parts discriminative enough for humans to recognize the class? 2) {\it Prototype recognition}: Can humans recognize the prototypes for parts of a certain image? 3) {\it Part-type likelihood prediction}: Do humans agree with the part-type likelihood scores output by the model?

\begin{figure}
\centering
\includegraphics[width=\columnwidth]{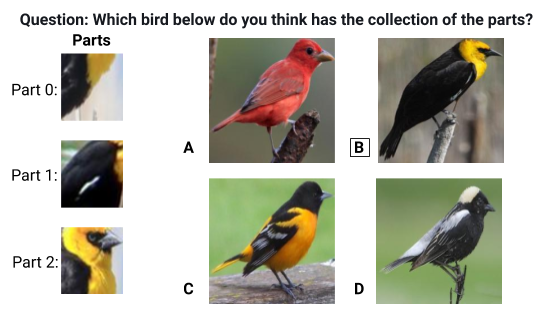}
\caption{\textbf{Q1.} Exemplar question for discriminability. The workers are asked to select the class they think has the parts provided on the left side. If none of the given classes applies, they should select ``None of above''. The ground truth class is B in this example which is bounded by a box (Best viewed in color).}
\label{fig:q1}
\end{figure}
\begin{table}[]
    \centering
    \begin{tabular}{l l}
    \hline
    \bf No. & \bf Class \\
    \hline
    0 & Yellow\_headed\_Blackbird*\\
    1 & Bobolink \\
    2 & Indigo\_Bunting \\
    3 & Painted\_Bunting* \\
    4 & Vermilion\_Flycatcher \\
    5 & American\_Goldfinch \\
    6 & Baltimore\_Oriole \\
    7 & Tree\_Swallow \\
    8 & Summer\_Tanager \\
    9 & Prothonotary\_Warbler\\
    \hline
    \end{tabular}
    \caption{Selected classes of CUB for crowd-sourcing. *: the selected unseen classes.}
    \label{tab:mturk_classes}
\end{table}

For the purpose of this section, we select a subset containing 10 classes of the CUB dataset. The classes are listed in Table~\ref{tab:mturk_classes}. Each class has around 60 images. To simulate the FSL and ZSL scenarios where the novel classes have no samples during training, we held out two as unseen classes, and train our model on the remaining 8. The model has 3 parts and the number of prototypes in each part is set to 5. After the model is trained, the 3 parts located by our model are cropped from the original images for all classes centered around the peak values of the attention maps. In each part, we select one example that represents each prototype through nearest neighbor search in the RPC features of the seen class images.

\begin{figure}[t]
\centering
\includegraphics[width=\columnwidth]{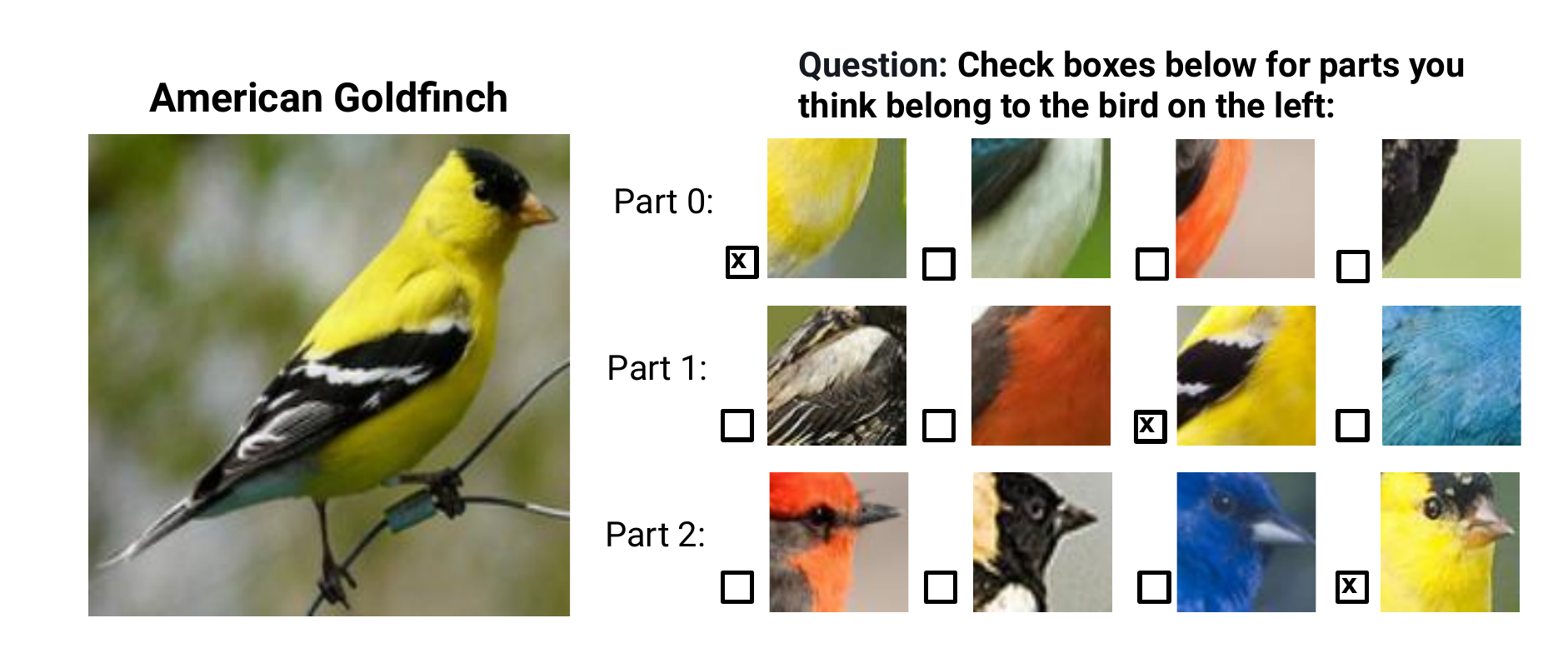}
\caption{\textbf{Q2.} Exemplar question for prototype recognition. Turkers are asked to pick the prototypes they think belong to the given image on the left. The ground truth is marked with an 'x' in its box (best viewed in color).}
\label{fig:q2}
\end{figure}

\begin{figure}
\centering
\includegraphics[width=\columnwidth]{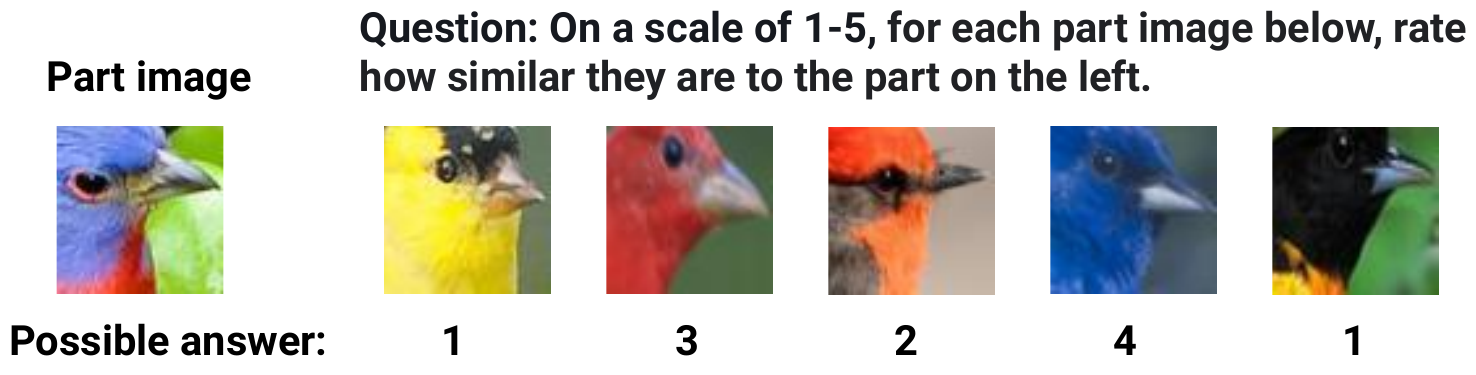}
\caption{\textbf{Q3.} Exemplar question for part-type likelihood prediction. Turkers are asked to choose a score for the similarity between the given part and the prototype. 1 means least and 5 means most similar (best viewed in color).}
\label{fig:q3}
\end{figure}

We use these prototypes and examples to create the assignments in MTurk. Each assignment is answered by 5 different turkers. The questions and results are detailed below:

\begin{figure*}[t!]
    \centering
     \begin{subfigure}[b]{0.49\textwidth}
         \centering
         \includegraphics[width=\textwidth]{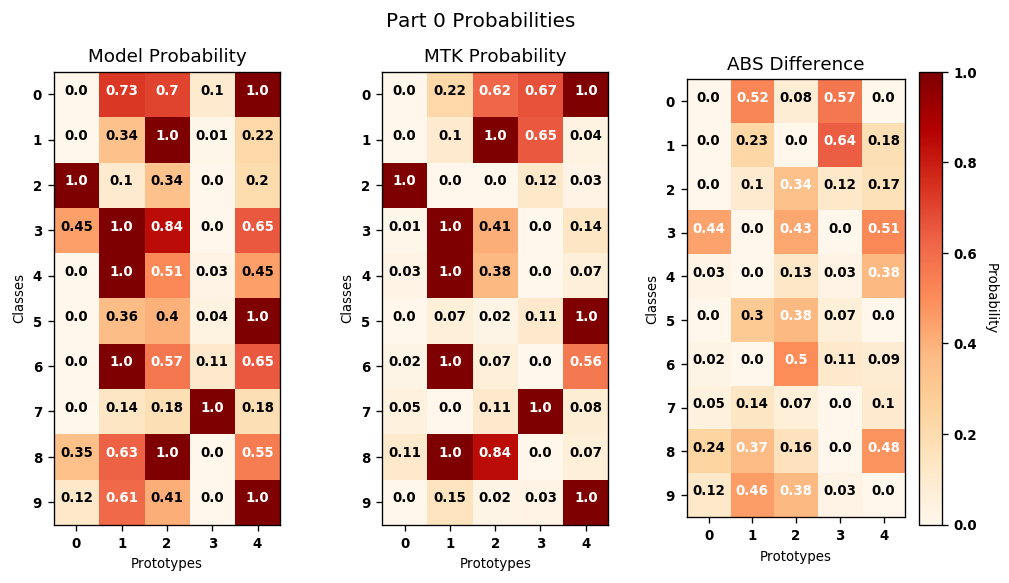}
         \caption{Part 0}
         \label{fig:part_0_prob}
     \end{subfigure}
     \hfill
     \centering
     \begin{subfigure}[b]{0.49\textwidth}
         \centering
         \includegraphics[width=\textwidth]{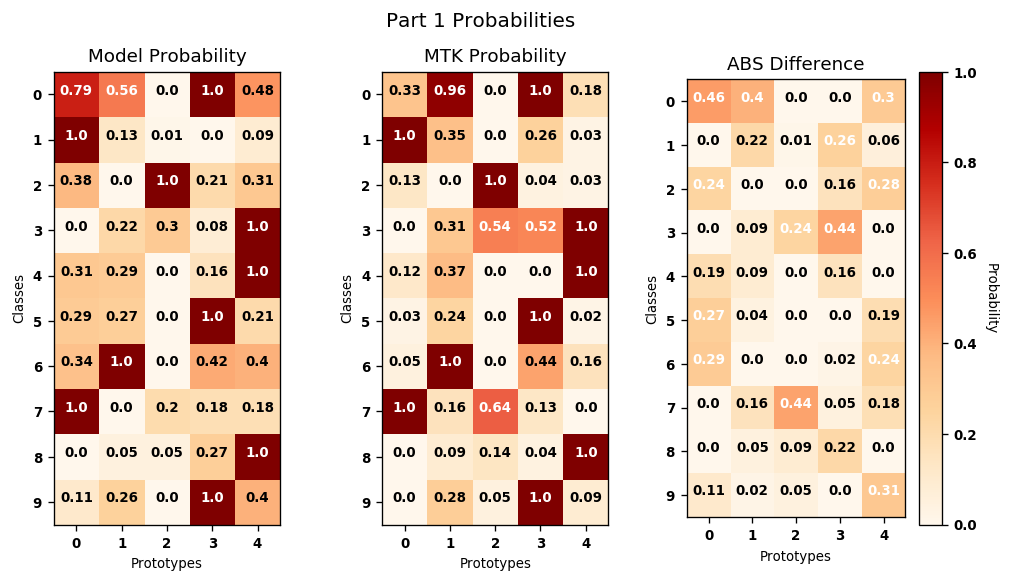}
         \caption{Part 1}
         \label{fig:part_1_prob}
     \end{subfigure}
     \hfill
     \begin{subfigure}[b]{0.56\textwidth}
         \centering
         \includegraphics[width=\textwidth]{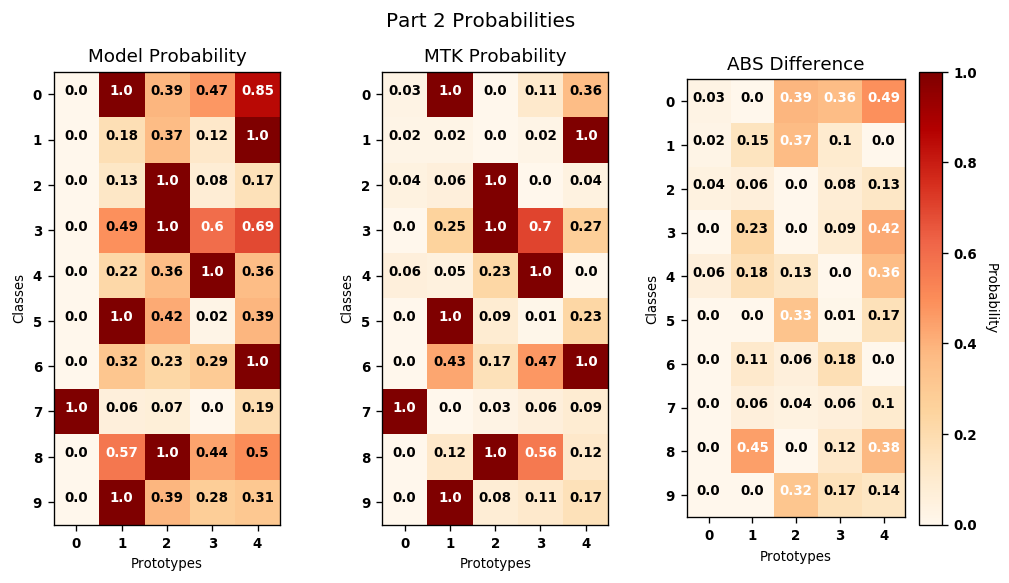}
         \caption{Part 2}
         \label{fig:part_2_prob}
     \end{subfigure}
     \vspace{-2mm}
     \caption{The class probability scores from our model (left) and the Mturk workers (middle), and their absolute difference (right). The RPC model's predictions of the likelihood that part instances come from specific concepts/prototypes agree with the predictions of humans}
     \vspace{-3mm}
\end{figure*}

\noindent\textbf{Discriminability of parts.} This question provides turkers with a collection of 3 parts identified by our model and asks them to select the class which those parts belong to. An exemplar question is shown in Fig.~\ref{fig:q1}. The turker can choose one class out of four options or ``None of the above''.
The intent of this question is to ascertain whether humans find the parts discriminative enough to distinguish between different categories, and demonstrates that our model learns to recognize salient parts of an image.
For this question we used 502 examples in total from all the 10 classes. We use the majority response of the 5 workers as the final answer for each image and consider the image to be ``correct'' if the final answer is the same as the ground truth class. The accuracy we obtained for all examples is 94.2\%, which indicates a high probability that humans can distinguish the bird using only the parts discovered by our model. Note that our model was only trained on the 8 seen classes, but it still produced discriminative parts for the novel classes, demonstrating that the parts learned are generalizable.

\noindent\textbf{Prototype recognition.} This question provides workers with an image, which we shall refer to as the query image, from one of the seen classes and a set of prototypical examples for each part. The workers are asked to choose the prototype in each part that they think belongs to the bird in the given image, as shown in Fig.~\ref{fig:q2}. The question is intended to determine if humans can learn to recognize images using the prototypes learned by our model. Good accuracy would demonstrate that the compact vocabulary of prototypes/concepts that our model learns, are meaningful to humans, and help them recognize whether a certain instance comes from a given learned concept.
We used 406 examples from seen classes for this question and in each part we also use the majority vote of the 5 answers for computing accuracy. For each prototype, we find the closest set of 5 examples as a representative set and the set of classes that these 5 images belong to is called the class-set corresponding to that prototype. Each question picks 4 prototypes randomly, and 1 out of the 5 representative examples is displayed on the question to represent that prototype. A response is considered ``correct'' if the class of the query image lies in the class-set of the prototype selected. 
The accuracy for the three parts are 92.96\%, 95.73\%, and 96.98\%, respectively, validating that humans can learn to recognize images in terms of concepts represented by the prototypes learned by our model. Note that a random guess would result in a 25\% accuracy in expectation. Also, note that the intention of this question is to evaluate the quality of our prototypes/concepts that form the vocabulary for encoding part instances in. It does not evaluate the agreement between RPC model's prediction of whether an image corresponds to a certain concept and a human's prediction. This is evaluated in the next question.

\begin{table*}[t]
    \centering
\renewcommand{\arraystretch}{1}
\setlength{\tabcolsep}{0.3cm}
\scalebox{1.0}{
    \begin{tabular}{l| c c c|c c c|c c c}
        \toprule
        \multirow{2}{*}{\bf{Methods}} & \multicolumn{3}{c|}{\bf{CUB}} & \multicolumn{3}{c|}{\bf{CUB-syn}} & \multicolumn{3}{c}{\bf{Cars-syn}} \\
         & U & S & H & U & S & H & U & S & H\\
        \midrule
        CADA-VAE \cite{schonfeld2018generalized} & \bf \color{red} 51.6 & 53.5 & \bf \color{red} 52.4 & \bf \color{red} 64.0 & 54.7 & \bf \color{red} 59.0 & \bf \color{red} 65.3 & \bf \color{red} 77.7 & \bf \color{red} 70.9 \\
        fCLSWGAN \cite{xian2018feature} & \bf \color{blue} 43.7 & 57.7 & \bf \color{blue} 49.7 & \bf \color{blue} 56.0 & \bf \color{blue} 59.5 & \bf \color{blue} 57.7 & \bf \color{blue} 46.1 & \bf \color{blue} 57.2 & \bf \color{blue} 51.1 \\
        RN \cite{sung2018learning}       & 38.1 & \bf \color{blue} 61.1 & 47.0 & 54.2 & \bf \color{red} 60.5 & 57.2 & 30.6 & 49.0 & 37.7 \\
        GDAN \cite{huang2018generative}     & 39.3 & \bf \color{red} 66.7 & 49.5 & 20.5 & 33.2 & 25.4 & 8.1  & 38.4 & 13.4 \\
        \bottomrule
    \end{tabular}}
    \caption{GZSL results on CUB, CUB-syn and Cars-syn. CUB uses original class semantic vectors and CUB-syn and Car-syn use synthetically generate class semantic vectors using the RPC model. U = unseen classes, S = seen classes, H = harmonic mean. The accuracy is class-average Top-1 in \%. In each column, \textcolor{red}{\bf red}=highest and \textcolor{blue}{\bf blue}=2$^{nd}$ highest accuracy}
    \label{tab:synthetic_attr}
\end{table*}

\noindent\textbf{Part-type likelihood prediction.} In this question, turkers are provided with a part example and a list of all prototypes for the same part. Turkers are asked to rate the probability on a scale of 1 to 5, 1 being least possible, that the given example belongs to a particular prototype, as shown in Fig.~\ref{fig:q3}. We used 502 images from all classes per part for this experiment. For each of the 3 parts, in order to obtain the class probability score, we average the scores provided by turkers 
over all images belonging to a certain category. As a comparison, we also average the RPC encoding from our model for each category. The class probability scores of the workers, our model, and their absolute differences, are visualized in Fig.~\ref{fig:part_0_prob}, Fig.~\ref{fig:part_1_prob} and Fig.~\ref{fig:part_2_prob} for the three parts, respectively. For better visualization, the probability scores are linearly scaled to lie in the range $[0, 1]$.
The results confirm that human annotators largely agree with the probability that our model assigns for a part in an image belonging to a given part-type represented by the prototypes. This holds true for scores from classes 0 and 3 as well, which are not seen during training. Thus the model's perception of visual similarity agrees with those of humans and this perception generalizes to unseen images.

\section{Synthetic Attribute Generation}
Learning at large-scale poses two fundamental challenges. First, acquiring ground-truth annotations for training instances is expensive. Second, as we scale the number of object classes, we observe few instances for many (rare) object classes. ZSL proposes to overcome these challenges by leveraging semantic descriptions or attributes for the different classes. 

In practice, advancements in ZSL methods is fundamentally limited by the unavailability of ``good'' zero-shot datasets. Semantic attributes that are visually well-aligned, such as in the CUB dataset, can be obtained through crowd-sourcing but doing so is often very expensive. Popular but less effective are datasets leveraging language corpora to derive encodings, such as word-embeddings, for different object class categories. Such word-embeddings although semantically meaningful, are visually misaligned, and may not represent a visually meaningful description of the classes.

Since our framework outputs image encodings that are interpretable by humans, we propose to leverage them as a proxy for human annotations to provide synthetic semantic attributes. Zero-shot learning methods can be evaluated using these attributes when human annotated attributes are unavailable. 

To generate synthetic semantic attributes, we train our RPC model for classification on all ``seen'' classes of the dataset. In this case, for our RPC model, we use $M=3$ parts and $K=64$ prototypes per part. Once trained, we generate RPC encodings of images from both seen and unseen classes, and use the average RPC encoding for each class, as its semantic vector. Note that this is similar to how the semantic class attributes were generated for the CUB dataset, the only difference is that attributes annotations per image were collected from humans for CUB. Note that we increased the number of prototypes used, compared to our previous experiments, to have a semantic vector that is similar to the size of the semantic vector for CUB (Our semantic vector size $=64 \times 3 = 192$ compared to the original $312$-dimensional semantic vectors in CUB).

Using the approach above, we generated semantic vectors for the CUB dataset \cite{WahCUB_200_2011}, which already has human annotated semantic vectors and the Cars dataset \cite{KrauseStarkDengFei-Fei_3DRR2013}, which is not a ZSL dataset and does not have any semantic vectors associated with class labels. We hence propose a novel GZSL split for the Cars dataset. Classes are split into 131 seen and 65 unseen. For each seen class, we randomly sample $3/4^{th}$ of its images as training data, and use the rest for testing (test-seen). All images of unseen classes belong to test-unseen. The  proposed  split  has 8,100 training images, 2,637 test-seen images and 5,448 test-unseen images. This split, along with the synthetically generated semantic vectors, will be made publicly accessible.

We report the result of 4 zero-shot learning methods (citations in table), which we evaluated using implementations from the authors' publicly available code. Results of the evaluation are reported in Table \ref{tab:synthetic_attr}. The columns titled CUB-syn and Cars-syn are the respective datasets with the synthetically generated semantic vectors. For this evaluation, we used the original hyperparameters of the specific GZSL methods for their evaluation on the CUB dataset. Note that we used the same hyperparameters for both CUB-syn and Cars-syn.

First, comparing the harmonic mean accuracies (H) for CUB and CUB-Syn in Table \ref{tab:synthetic_attr}, we see that all methods perform similarly as they did on the original semantic attributes on CUB, with CADA-VAE performing the best. All methods improve in absolute accuracy, possibly since our attributes better mirror visual information than the semantic attributes that are collected via a noisy crowd-sourcing procedure. This improvement is with the exception of GDAN, which decreases in performance, possibly due to its sensitivity to hyperparameters, corresponding to the specific semantic vectors. On the Cars-syn GZSL dataset, using our 
synthetically generated class semantic vectors, we see that CADA-VAE and fCLSWGAN remain the two best performing methods, but it seems RN and GDAN do not do well because of their sensitivity to specific hyperparameters used.

\section{Conclusion}
We proposed Recognition as Part Composition, an approach for image recognition inspired by human cognition. Our approach first decomposes an image into salient parts, and then learns to represent each part instance as a mixture of a few concepts. We found that this approach, imparted big benefits to classifiers in low-shot generalization tasks like zero-shot learning, few-shot learning and unsupervised domain adaptation. We also found that using these encodings also makes a classifier more robust to adversarial attacks, which impercetibly change an input image to induce a classifier error. Via crowd-sourcing, we also demonstrated that the encodings agree with human perception and that humans can recognize images using the parts learnt by our model and recognize the part instances in the vocabulary that RPC learns. Given the fact our encodings are human-interpretable, we proposed an application of them, to generate synthetic attributes for evaluating zero-shot learning methods on new datasets, before collecting human annotated class semantics for them. We demonstrated this on the Stanford Cars dataset as a proof of concept.

\section*{Acknowledgements}

This research was supported by the Army Research Office Grant W911NF2110246,
the National Science Foundation grants CCF-2007350 and CCF-1955981, and the Hariri
Data Science Faculty Fellowship Grants. 
The authors would like to thank Ruizhao Zhu for helpful discussions.

\ifCLASSOPTIONcaptionsoff
  \newpage
\fi

\bibliographystyle{IEEEtran}
\bibliography{submission}

\begin{IEEEbiography}[{\includegraphics[width=1in,height=1.25in,clip,keepaspectratio]{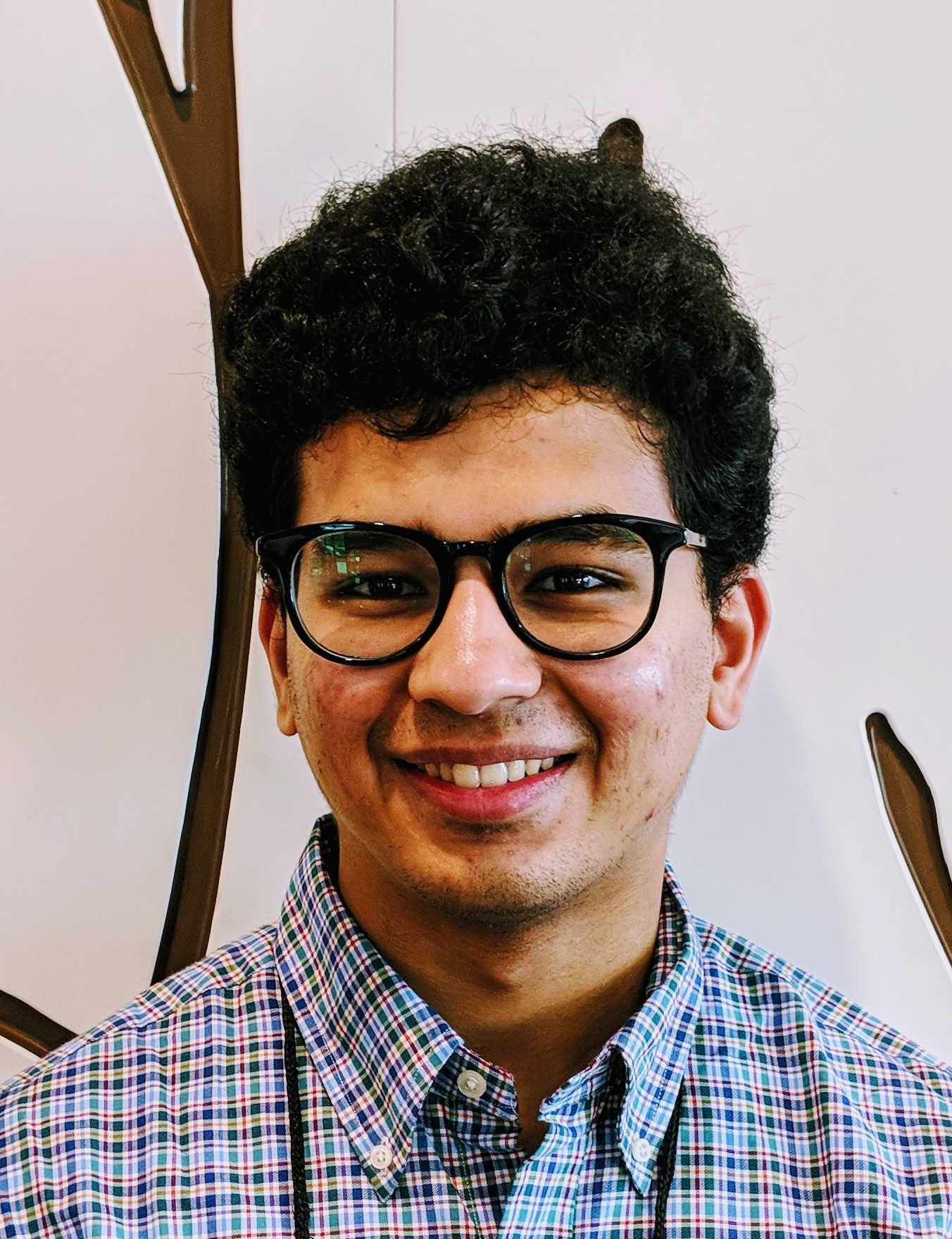}}]{Samarth Mishra}
is a Ph.D. student in the Image and Video Computing group at Boston University, co-supervised by Profs. Venkatesh Saligrama and Kate Saenko. His research interests lie in computer vision and machine learning, and specifically in problems dealing with a scarcity of labeled data. More information about his work can be found at \href{https://samarth4149.github.io}{https://samarth4149.github.io}
\end{IEEEbiography}

\begin{IEEEbiography}[{\includegraphics[width=1in,height=1.25in,clip,keepaspectratio]{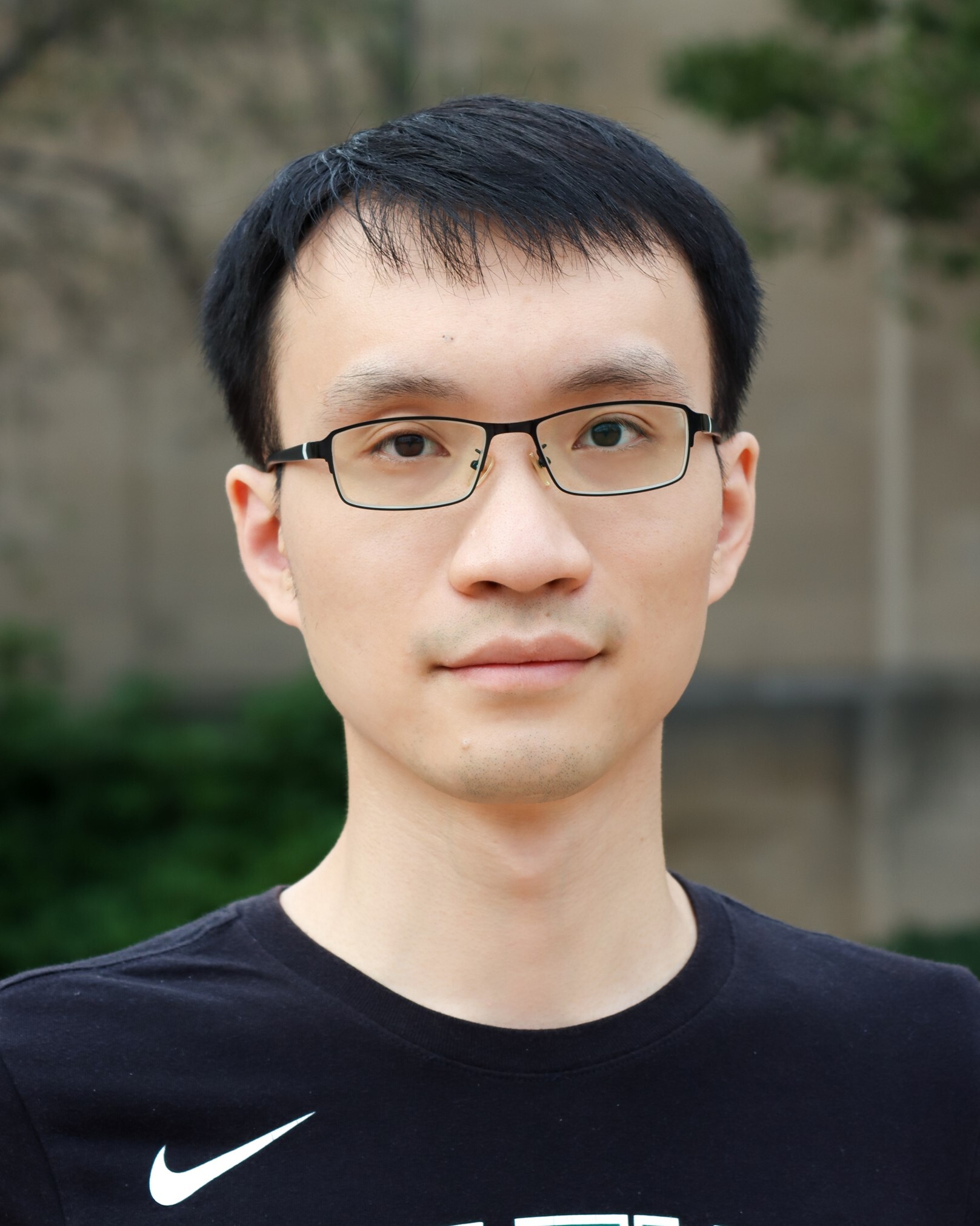}}]{Pengkai Zhu}
is an applied scientist in AWS AI. He received his Ph.D from Boston University, where he was supervised by Prof. Venkatesh Saligrama. His research interests include computer vision and machine learning, specially low-shot recognition and visual document understanding.
\end{IEEEbiography}

\begin{IEEEbiography}[{\includegraphics[width=1in,height=1.25in,clip,keepaspectratio]{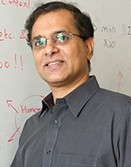}}]{Venkatesh Saligrama} 
is a faculty member in the Department of Electrical and Computer Engineering, the Department of Computer Science (by courtesy), and a founding member of the Faculty of Computing and Data Sciences at Boston University. His research interests are broadly in the area of Artificial Intelligence, and his recent work has focused on machine learning with resource-constraints. He is an IEEE Fellow and recipient of several awards including Distinguished Lecturer for IEEE Signal Processing Society, the US Presidential Early Career Award (PECASE), ONR Young Investigator Award, and the NSF Career Award. More information about his work is available at \href{http://sites.bu.edu/data}{http://sites.bu.edu/data}
\end{IEEEbiography}

\vfill

\end{document}